\documentclass[10pt,conference,final]{IEEEtran}
\usepackage{cite}
\usepackage{amsmath,amssymb,amsfonts}
\usepackage{algorithmic}
\usepackage{graphicx}
\usepackage{textcomp}
\usepackage{todonotes}
\usepackage{hyperref}
\usepackage{wrapfig,lipsum,booktabs}
\usepackage{multirow}
\usepackage{stackengine}
\usepackage{subcaption}
\usepackage{caption}
\usepackage{float}
\usepackage{abstract}
\usepackage[noblocks]{authblk}

\def\BibTeX{{\rm B\kern-.05em{\sc i\kern-.025em b}\kern-.08em
    T\kern-.1667em\lower.7ex\hbox{E}\kern-.125emX}}
\begin{document}

\title{DEEMD: \underline{D}rug \underline{E}fficacy \underline{E}stimation against SARS-CoV-2 based on cell \underline{M}orphology with \underline{D}eep multiple instance learning}

\author{M.Sadegh Saberian, Kathleen P. Moriarty, Andrea D. Olmstead, Christian Hallgrimson, \\François Jean, Ivan R. Nabi, Maxwell W. Libbrecht, Ghassan Hamarneh$^{*}$, \textit{Senior Member, IEEE}.

\thanks{
{This work was supported by operating grants from the Canadian Institutes of Health Research (CIHR) / Operating Grant: COVID-19 May 2020 Rapid Research Funding Opportunity [VR3-172639 (F.J., I.R.N., G.H.)], the Coronavirus Variants Rapid Response Network (CoVaRR-Net) [CIHR 175622 (F.J., A.O., I.R.N.)], and Natural Sciences and Engineering Research Council of Canada (NSERC) Alliance COVID-19 grant: ALLRP 553515-20 (to G.H., F.J., and I.R.N.).}}
\thanks{M.Sadegh Saberian, Kathleen P. Moriarty, Christian Hallgrimson, Maxwell W. Libbrecht, and Ghassan Hamarneh are with School of Computing Science, Simon Fraser University, BC, Canada, V5A 1S6. e-mail: \{msaberia, kmoriart, cdh13, maxwl, hamarneh\}@sfu.ca}
\thanks{Andrea D. Olmstead and François Jean are with Department of Microbiology and Immunology, Life Sciences Institute, University of British Columbia, BC, Canada, V6T 1Z3. email: andrea.olmstead@ubc.ca, fjean@mail.ubc.ca}
\thanks{Ivan R. Nabi is with Department of Cellular and Physiological Sciences, School of Biomedical Engineering, Life Sciences Institute, University of British Columbia, BC, Canada, V6T 1Z3. e-mail: irnabi@mail.ubc.ca}
\thanks{Correspondence to: Ghassan Hamarneh \{\textit{hamarneh@sfu.ca}\}}
}

\maketitle

\begin{abstract}
Drug repurposing can accelerate the identification of effective compounds for clinical use against SARS-CoV-2, with the advantage of pre-existing clinical safety data and an established supply chain. RNA viruses such as SARS-CoV-2 manipulate cellular pathways and induce reorganization of subcellular structures to support their life cycle. These morphological changes can be quantified using bioimaging techniques. In this work, we developed DEEMD: a computational pipeline using deep neural network models within a multiple instance learning framework, to identify putative treatments effective against SARS-CoV-2 based on morphological analysis of the publicly available RxRx19a dataset. This dataset consists of fluorescence microscopy images of SARS-CoV-2 non-infected cells and infected cells, with and without drug treatment. DEEMD first extracts discriminative morphological features to generate cell morphological profiles from the non-infected and infected cells. These morphological profiles are then used in a statistical model to estimate the applied treatment efficacy on infected cells based on similarities to non-infected cells. DEEMD is capable of localizing infected cells via weak supervision without any expensive pixel-level annotations. DEEMD identifies known SARS-CoV-2 inhibitors, such as \textit{Remdesivir} and \textit{Aloxistatin}, supporting the validity of our approach. {DEEMD can be explored for use on other emerging viruses and datasets to rapidly identify candidate antiviral treatments in the future}. Our implementation is available {online at {\url{https://www.github.com/Sadegh-Saberian/DEEMD}}}.
\end{abstract}

\begin{IEEEkeywords}
Drug Repurposing, Deep Multiple Instance Learning, Morphological Analysis, SARS-CoV-2. 
\end{IEEEkeywords}
\saythanks
\section{Introduction}
\label{sec:introduction}
The COVID-19 global pandemic has urged the research community to focus their resources towards studying SARS-CoV-2 and discovering or identifying potential therapeutics. To date, despite intense efforts, very few treatment options are available for those suffering from COVID-19  \cite{TherapeuticManagement}. Drug repurposing is an attempt to identify existing clinically approved treatments with established pharmacological and safety profiles that could be rapidly redirected towards clinical treatment of novel diseases such as COVID-19  \cite{pushpakom2019drug,yousefi2020repurposing} (and references within). The antiviral activity of candidate compounds can be tested using cell-based systems of viral infection. The detection of the viral infection is achieved using molecular tools such as antibodies directed at virus-encoded proteins. In the case of newly emerging viruses such as SARS-CoV-2, access to such molecular tools may represent an important limiting step to rapidly developing cell-based assays to discover novel antiviral agents. Alternatively, since human pathogenic viruses such as SARS-CoV-2 manipulate cellular pathways to reorganize the host cell morphology to support their life cycle  \cite{long2020super,cortese2020integrative}, developing a computational method to perform quantitative analysis of virus-induced cell morphology provides a unique approach to discover candidate antiviral molecules without the use of viral biomarkers \cite{bray2016cell}. Virus infected cells can be treated with thousands of compounds at different concentrations followed by staining of cellular structures with fluorescent dyes that can be imaged using high-content screening microscopes  \cite{pepperkok2006high,starkuviene2007potential}. Morphological features can be extracted from images of infected and non-infected cells and then applied to images of infected, drug-treated cells to predict antiviral efficacy based on cellular morphology. Thus, quantitative morphological analysis of cells as a computational method for drug repurposing allows for accelerated parallel screening of multiple therapeutics  \cite{mirabelli2020morphological}.

\begin{figure*}[h!]
\centering
\centering
\includegraphics[width=\linewidth]{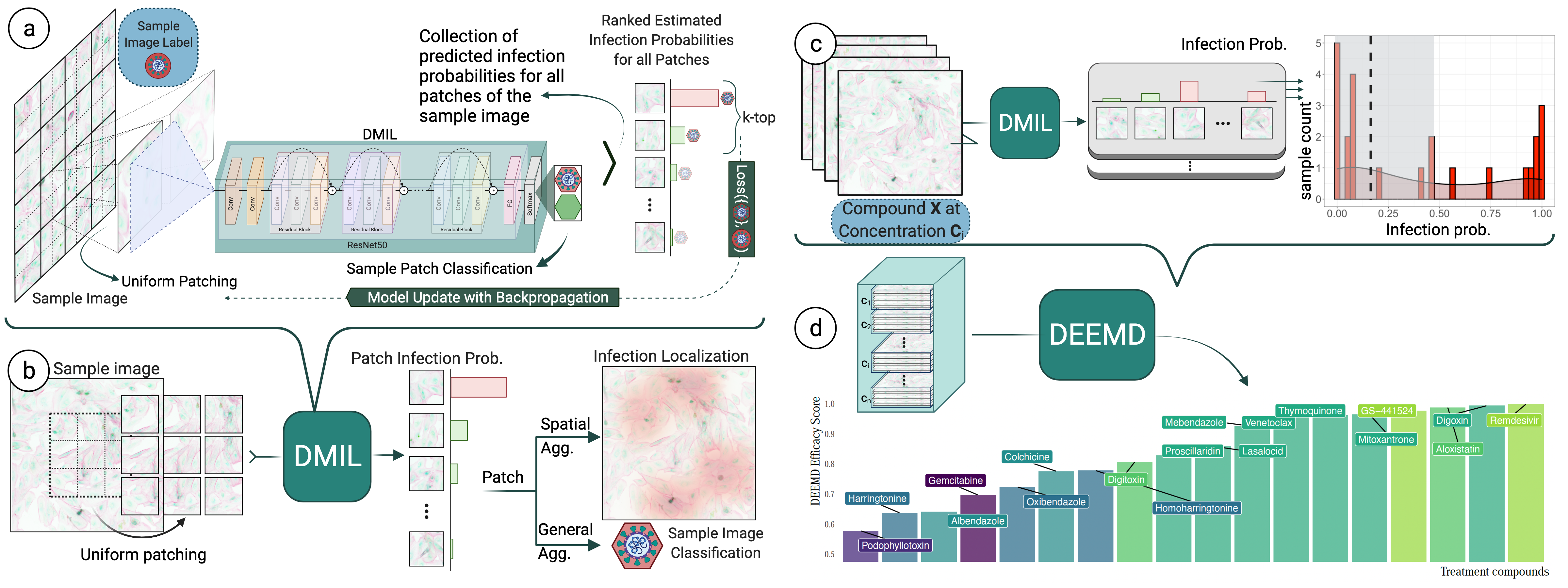}
\vspace{-5mm}
\caption{{Overview of the proposed pipeline: DEEMD \textbf{(a)} Training a deep neural network within a multiple instance learning framework (DMIL). \textbf{(b)} Generating infection maps and sample image classification based on the DMIL estimated patch infection probabilities. \textbf{(c)} Aggregating replicate experiments using non-parametric testing to estimate dose-dependent efficacy scores. \textbf{(d)} DEEMD estimates the efficacy of a treatment based on its multiple replicates and concentrations by combining deep learning and statistical modeling.}}
    \label{fig:overview}
\end{figure*}

{In this work, we present DEEMD: a computational pipeline capable of estimating and ranking the efficacy of treatments based on the morphological features of cells being treated. These morphological features are extracted using a deep learning model trained within a multiple instance learning (MIL) framework \cite{carbonneau2018multiple}. DEEMD is able to process thousands of treatments and can accelerate the identification of clinically evaluated and {US Food and Drug Administration (FDA) approved} compounds with antiviral activity against a pathogen agent. To show capabilities of this pipeline, we applied DEEMD on a public microscopy image dataset of SARS-CoV-2 infected cells, and it is able to identify treatment compounds that have been suggested in the literature to be effective against COVID-19.}

{An abstract overview of the proposed pipeline is shown in Fig~\ref{fig:overview}. The main underlying idea for DEEMD is that the effective compounds are those that are able to prevent infection from spreading and maintain the treated cells morphological features similar to the healthy samples. Based on this assumption, we train a deep neural network classifier (DMIL) to distinguish between the healthy and the infected sample images using weakly supervised training and MIL, Fig~\ref{fig:overview}-(a). The trained DMIL estimates an infection probability for each patch of a sample image. These infection probabilities are aggregated spatially into infection maps and globally to predict a label for unseen sample images during inference, Fig~\ref{fig:overview}-(b). The infection probabilities for treated samples of a specific compound at a defined dose are further analysed to calculate a dose-dependent efficacy estimation, Fig~\ref{fig:overview}-(c). The {treatment} dose-dependent efficacy scores are further aggregated into an efficacy score to rank and identify the effective ones, Fig~\ref{fig:overview}-(d).}

{The rest of the} paper is structured as follows: In Section~\ref{sec:background} we review the literature on drug repurposing, high-throughput cell imaging, and computational multiple instance learning framework. In Section~\ref{sec:method} we present DEEMD and its components and discuss them in detail. In Section~\ref{section:experimental} we describe the RxRx19a dataset and DEEMD implementation details. In Section~\ref{sec:res} we discuss our findings along with DEEMD's limitations. Finally, in Section~\ref{sec:conclusion} we conclude the paper.

\section{Background}\label{sec:background}
\subsection{Drug Repurposing and High-throughput Imaging}\label{sec:drugRepurposing}
Drug repurposing is an active research area in the pharmaceutical industry in which existing drugs are used in alternative applications from the original design and it has been shown to be an effective approach in many cases \cite{pushpakom2019drug}. 
{For instance, \textit{Fingolimod} was originally developed to prolong organ graft survival. But later through drug repurposing, it was found to be effective against multiple sclerosis pathogenic immune responses  \cite{brinkmann2010fingolimod}.} 
Various approaches are used to identify candidate drugs for repurposing applications including computational methods, i.e. genetic association or molecular docking, and experimental methods, i.e. phenotypic screening or binding assays. An additional approach is to apply high-throughput imaging assays for {image-based cell profiling or morphological profiling} to predict antiviral effectiveness. { It has been shown that cell morphology is an observable, measurable signal of particular cellular states or cell processes; therefore, it has the potential to be used in analysis of cells interaction with perturbations (genetic or chemical) as an unbiased source of information \cite{ziegler2021morphological,caicedo2017data}. However, not all of these morphological changes and alterations are induced by the introduced perturbations. In fact, many of them are not even obvious to the human eye. Thus, there is a need for developing such analysis methods that are not only capable of capturing such morphological changes, but also able to discriminate the perturbation-induced ones. Throughout the paper, we use the term morphology to refer to the broad range of observable, measurable, biological phenotypes in images that includes shape, intensity, texture, spatial and stain-specific context features. {Some cell image analysis software}, such as CellProfiler \cite{mcquin2018cellprofiler}, provide a platform for the user to define a pipeline consisting of a diverse set for pre-defined classical metrics and features to be extracted from the images. On the other hand, with the substantial success of deep learning models, especially convolutional neural networks (CNN), in various fields of computer vision \cite{lecun2015deep}, these pipelines are also used for extracting the morphology features of the cells with promising results \cite{eulenberg2016deep,pawlowski2016automating,godinez2017multi,long2020super,sommer2017deep}. It should be noted that the extracted features from the CNNs are not fundamentally different to the classical ones. It has been shown that the filters in CNNs learn to extract progressively abstract features from the input images. In fact, correspondence between the learned feature maps and the classical features such as edges and textures has been identified \cite{zeiler2014visualizing}. Morphological profile refers to the set of features, classical or extracted by a trained deep neural network, describing a sample cell population. These features can be leveraged to quantitatively compare the morphological changes induced by each compound in a cell population. These morphological profiles can also provide information about how compounds may be interacting with the host cell, their molecular targets, and the affected cellular pathways. Image-based cell profiling  has been successfully utilized in small molecule profiling  \cite{bray2016cell,wawer2014toward}, identifying drug mechanism of action  \cite{kraus2016classifying,cox2020tales,subramanian2017next}, and in drug repurposing  \cite{simm2018repurposing,mirabelli2020morphological}.}

    Recently, many drug repurposing studies have focused on clinically evaluated and {FDA-approved} therapeutic compounds, to find treatments that exhibit antiviral activity against SARS-CoV-2  \cite{zhou2020artificial,kuleshov2020covid}. Among multiple methodologies, high-throughput screens have enabled image-based morphological analysis of cells infected with SARS-CoV-2   \cite{mirabelli2020morphological,heiser2020identification,cuccarese2020functional,pham2021deep}. Mirabelli \textit{et al.}  \cite{mirabelli2020morphological} used Huh-7 cells infected with SARS-CoV-2 and treated with a library of 1425 FDA-approved compounds at different concentrations to identify compounds with antiviral activity. The samples were then probed for SARS-CoV-2 nucleocapsid protein and dyes specific to cell organelles for imaging. CellProfiler  \cite{mcquin2018cellprofiler} was used to extract morphological features from the infected cells, using SARS-CoV-2 nucleocapsid as an indicator for regions of interest, and trained a random forest model to predict efficacy scores for each treatment. These scores were used to select efficacious compounds for follow-up experimental triplicate dose-response confirmation. They identified 17 effective compounds including \textit{Remdesivir}, of which 10 are novel \textit{in vitro} identifications.

Heiser \textit{et al.}  \cite{heiser2020identification} used a morphological analysis pipeline on the RxRx19a dataset, a fluorescence  microscopy dataset of Human Renal Cortical Epithelial (HRCE) cells, to identify potential treatments for SARS-CoV-2. They used a proprietary deep convolutional neural network to embed the sample images for calculating on-disease projection and off-disease rejection scores per dose for each treatment. These scores were then aggregated and normalized using a proprietary algorithm to compute the hit-scores. They reported that \textit{Remdesivir}, \textit{GS-441524}, \textit{Aloxistatin}, \textit{Silmitasertib}, and \textit{Almitrine} showed moderate to strong effectiveness against SARS-CoV-2 in their model, whereas, neither \textit{chloroquine} nor \textit{hydroxychloroquine} demonstrated any significant antiviral effectiveness. Similarly, Cuccarese \textit{et al.} \cite{cuccarese2020functional} used an \textit{in vitro} deep-learning-driven analysis of cellular morphology on Human Umbilical Vein Endothelial Cells (HUVEC) treated with protein cocktails that mirror those from severe COVID-19 to identify potential drug repurposing candidates against COVID-19-associated cytokine storm.

\subsection{Multiple Instance Learning}\label{sec:MIL}
Multiple instance learning (MIL), as a form of weakly supervised learning, has been under the spotlight of research communities recently due to its ability to leverage weak supervision for tasks that are conventionally considered heavily dependent on laborious human annotations  \cite{carbonneau2018multiple}. MIL was originally introduced for drug activity prediction  \cite{dietterich1997solving} and recently it has been applied in many different domains such as computer vision  \cite{wan2019c}, medical imaging and diagnosis  \cite{yao2020whole}, and in biology for applications such as mechanism of action classification using microscopy images  \cite{kraus2016classifying}, identifying antigen binding peptides  \cite{yasser2010predicting}, and predicting specific functional binding sites in microRNA targets  \cite{bandyopadhyay2015mbstar}. Deep multiple instance learning uses different MIL approaches combined with a deep neural network model as a learner in the MIL framework, and it has shown performance competitive with state-of-the-art in recent studies \cite{campanella2019clinical,ilse2018attention,kraus2016classifying,yan2018deep,zhao2020predicting}.

Contrary to conventional supervised learning methods, where every instance in the training data is associated with a label, in the MIL framework the learner is provided with a training set of labeled bags, where each bag is a set of instances. The learner is tasked to predict a label for an unseen bag given its instances. {Instance-space} algorithms are a category of MIL which assume that the discriminative information lies within instances and is local. Hence, they predict the bag label by training an instance classifier and then aggregating the instance-level predictions into the bag label \cite{amores2013multiple}. The most commonly used assumption in these methods -- the standard multiple instance assumption -- states that in a binary bag classification setup, each bag labeled with the positive label contains at least one instance representing the positive class, whereas the negative class bags do not contain any instances of positive class.

Virus-specific markers are not always included in high-throughput imaging studies that investigate the effects of viral infection and drug treatments on host cell morphology; in these instances, the extent of viral infection is unknown. This scenario may also occur during the very early stages of an outbreak or pandemic when limited virus-specific tools are available, such as when COVID-19 first emerged. When no pixel-level annotations are available to specifically identify infected regions of the imaged cell population, MIL may be used as an alternative strategy to localize viral infection and the corresponding effects on host cell morphology. The imaged cell populations can be broken down into smaller instances, single cells or patches, and the sample cell population label is utilized in the MIL formulation. An instance-space approach would suit this problem very well for viral infections, such as SARS-CoV-2, that induce local morphological changes to cells  \cite{cortese2020integrative}. The standard assumption is that if a sample cell population is infected, then there should be at least one instance that contains infected cells, and a non-infected sample cell population does not have any infected cells.

\subsection{Contributions}\label{sec:contributions}

To the best of our knowledge, this is the first work to approach the drug repurposing problem via leveraging weak supervision through multiple instance learning. Some morphological analysis pipelines utilize a viral-specific marker to localize infected regions of interest in sample images for studying morphological changes induced by the virus, such as \cite{mirabelli2020morphological}. Another approach is to rely on computer vision models to extract viral-related morphological features through massive datasets without providing any annotations on infected regions of interest \cite{heiser2020identification}. DEEMD integrates both these approaches through identifying infected regions in sample images without requiring any viral-specific markers or annotations. DEEMD can be applied to quantify the effectiveness of any available compound and identify lead candidates that show therapeutic potential against viral infection. The identified candidates can then be evaluated using other molecular tools once they become available, such as fluorescent-tagged viruses, nucleic acid-based methods (PCR), or animal models.

\section{Methodology}\label{sec:method}

We first describe a high-level overview of DEEMD before delving into the details. DEEMD consists of two components: the first component is a deep learning model which is trained on microscopy images of untreated samples to extract discriminative morphological features in MIL framework to distinguish between uninfected and SARS-CoV-2-infected cell population images, further described in Section~\ref{Sec:mil_classification}. Following infection of cells treated with a drug, the trained deep learning model is used to extract a morphological {profile for each patch in the sample images.} These profiles are then aggregated and used in a statistical model, the second component, to estimate the treatment efficacy per concentration, Section~\ref{Sec:EfficacyEstimation}. Finally, all dose-dependent efficacy  scores are aggregated for each treatment to form the final identified set of potentially effective treatments. In this context, we define treatment effectiveness as how close a treated sample's morphological profile is to uninfected cell profiles. We discuss each component in detail in the following sections. For brevity, {Table~\ref{notations-tab:1} in the Section~\ref{appendix} (Appendix)} includes all notations used in Section~\ref{sec:method}.

\subsection{Classification with deep MIL}\label{Sec:mil_classification}

In this work, we follow closely the training procedure described in \cite{campanella2019clinical,yan2018deep} for training an instance classifier based on a relaxed version of the standard multiple instance assumption. The training set $\mathcal{D}$ consists of $n$ data points $(\mathbf{X},\mathbf{Y})$ in form of $(x_i,y_i)$ pairs where $x_i$ is {an image from the set of} fluorescence microscopy {images  $\mathbf{X}$}, along with its associated sample-level label $y_i$ {from set $\mathbf{Y}$} with {binary} classes: SARS-CoV-2 infected ({$1$}) and non-infected ({$0$}). Each sample image $x_i$ can be split into $N$ patches each referred to as $x_i^{p_{j}}$. We assume that each $x_i^{p_{j}}$ is associated with a patch-level label $y_{i}^{p_j}$ which is unknown and not included in the training set $\mathcal{D}$. Let 
$\mathcal{M}(\theta)$ be a deep neural network model {with parameters $\theta$, which is }responsible for {providing} a patch-level {infection probability} ${\mu}_i^{p_j}$ for every patch $x_i^{p_j}$ of each sample image $x_i$. In the MIL terminology, model $\mathcal{M}(\theta)$ is an instance classifier since it classifies each instance into SARS-CoV-2 infected and non-infected classes. We define $\mu_{i}^{p_j}$ to be the {MIL} model $\mathcal{M}(\theta)$ estimation of the unknown true patch-level label $y_{i}^{p_j}$ given the dataset $\mathcal{D}$. This can be expressed mathematically as:
\begin{equation}
     \mu_{i}^{p_j} =  \mathbb{P}\Big[y_i^{p_j} = {1} \Big | \mathcal{D};\mathcal{M}(\theta) \Big],
\end{equation}
where $\mathbb{P}[A]$ shows the probability of the event $A$.

Each iteration within a training epoch starts with an exhaustive inference using the model from previous iteration over all patches in the training set and all corresponding $\mu_{i}^{p_j}$ are estimated. Then for each sample image $x_i$, all of its patch-level {infection probability predictions} $\mu_{i}^{p_j}$ are sorted and the set of $k$ patches with the highest infection probability, denoted as $\mathcal{K}_{i}^k$, are selected for model training. 
The sample-level label $y_i$ is assigned {to each patch} in the set $\mathcal{K}_{i}^k$, and the {MIL} model $\mathcal{M}(\theta)$ is trained to minimize the binary cross entropy loss, $\mathcal{L}(\cdot)$, between $y_i$ and $\mu_{i}^{p_j}$ for all patches in the set $\mathcal{K}_{i}^k$. 
\begin{multline}\label{eq:loss}
\mathcal{L}\big(x_i |\mathcal{D};\mathcal{M}(\theta),w^+,w^-\big) =\\-\frac{1}{k}\displaystyle\sum_{j\in \mathcal{K}_{i}^k} \Big[  w^+y_i \text{log}(\mu_{i}^{p_j}) +  w^-(1-y_i)\text{log}(1 -\mu_{i}^{p_j})\Big],
\end{multline}
where $w^+$ and $w^-$ are incorporated in the loss function $\mathcal{L}(\cdot)$ {as real number weights} to account for class imbalance \cite{king2001logistic}. {Finally, the value of the loss function is averaged over the whole dataset in each epoch and back-propagates to update the model parameters.} For details on how to tune hyper-parameter $k$ see {Section~\ref{k-selection}}. The reader is encouraged to refer to {the Section~\ref{supp:noise}} for a detailed discussion on how hyper-parameter $k$ affects the training dynamics for a MIL learner.

\subsection{Inference with deep MIL}\label{sec:MILinference}
\begin{sloppypar}
Since the classification labels are only available for the sample images, aggregation of the patch-level infection probabilities is required for performance evaluation and downstream analysis. For inference, each sample image {$x_i$ is split into patches and each patch is passed forward through the model. This provides us with the estimated infection probabilities for all patches of the sample image $x_i$, which are aggregated to form the} sample-level label $\hat{y}_i$:
{\begin{equation} \label{kSMI}
    \hat{y}_{i} = \mathbb{I} \Big({\max_j}\{\mu_{i}^{p_j}\} \geq \eta\Big), 
\end{equation}}
\noindent where $\eta$ is the cut-off threshold, selected based on the validation set and $\mathbb{I}(\cdot)$ is an indicator function {returning 1 if the boolean expression argument is true and 0 otherwise}.
\end{sloppypar}
For the MIL model $\mathcal{M}(\theta)$, hyper-parameter $k$ is incorporated into the standard multiple instance assumption {during the training phase} so that the model can generalize to a more relaxed constraint: a positive (SARS-CoV-2 infected) sample image should contain at least $k$ positive patches to be considered positive and a sample image would be considered negative if less than $k$ patches are predicted to be infected. {However, during the evaluation phase, validation or testing, we classify a sample image as positive if there is at least $1$ patch identified as positive by the model. This would make the comparison between multiple choices of $k$ fair and prevents biases introduced by using a fixed value of $k$ for each sample during the training.}

\subsection{Treatment Efficacy Estimation}\label{Sec:EfficacyEstimation}
DEEMD estimates the treatment efficacy based on its trained MIL model that compares the morphological profile of drug-treated cells with those of untreated infected and uninfected cells. We assume that an effective treatment would prevent drastic infection-induced morphology changes in the cells, hence the treated cells morphological profile would be similar to that of uninfected ones. A statistical model takes in the infection probabilities predicted by the MIL model and estimates the probability that a treatment is effective based on its morphological similarities to uninfected cells. For each sample image $x_i$, the infection probability $z_i$ is calculated as:
\begin{equation}
    z_i = \text{median} \{\mu_i^{p_j}|p_j \in \mathcal{K}_{i}^k\}.
\end{equation}
We define the set $\mathcal{T}_{t_i}^{c_j}$ such that it consists of infection probabilities for every sample image of a given treatment $t_i$ at concentration $c_j$ in the treated test set. We have to aggregate the infection probabilities, $z_i$, over the set $\mathcal{T}_{t_i}^{c_j}$ to minimize the inevitable variations that have occurred during the sample preparation, treatment administration, and image acquisition. We define $e_{t_i}^{c_j}$ to be the dose-dependent estimated efficacy score of the treatment $t_i$ at concentration $c_j$ based on any model $\Omega(\omega)$ capable of assigning an infection probability to a sample image based {on} its morphology. Since we do not have any assumptions on the distribution of infection probabilities $z_i$ in the set $\mathcal{T}_{t_i}^{c_j}$, we opt to use non-parametric statistics related to the median of this distribution, $\beta_{t_i}^{c_j}$. We observed that the distribution infection probabilities $z_i$ in the set $\mathcal{T}_{t_i}^{c_j}$ for any $t_i$ is heavily skewed and asymmetric, thus using Wilcoxon test is misleading since it assumes a symmetric distribution for the data\cite{wilcoxon1992individual}. However, the sign test is still valid\cite{dixon1946statistical}. Given a confidence level $a$, we can calculate the exact confidence interval for the estimated median, denoted as $\text{CI}(\beta_{t_i}^{c_j}|a,\Omega(\omega))$. To be more conservative about the false positive rate, instead of using the point estimate $\beta_{t_i}^{c_j}$, we use the least upper bound, supremum, of $\text{CI}(\beta_{t_i}^{c_j}|a,\Omega(\omega))$. The mathematical expression for the dose-dependent estimated efficacy score of the treatment $t_i$ at concentration $c_j$ is:
\begin{equation}
    e_{t_i}^{c_j} = T(\mathcal{T}_{t_i}^{c_j}|\Omega(\omega))  = 1 - \text{sup}\big\{\text{CI}\big(\beta_{t_i}^{c_j}|a,\Omega(\omega)\big)\big\},
\end{equation}
where $T(\cdot)$ is a descriptive statistic. The dose-dependent efficacy score $e_{t_i}^{c_j}$ reflects the model $\Omega(\omega)$ belief on the morphological similarity between the sample images in the set {$\mathcal{T}_{t_i}^{c_i}$} and the uninfected ones in the training set. Close to $1$ values for $e_{t_i}^{c_j}$ imply that the treatment $t_i$ at concentration $c_j$ is effective against SARS-CoV-2. In this context, we assume that the morphological profile of cells treated with an effective treatment is more similar to the uninfected cells than infected ones. 

To summarise, all estimated dose-dependant efficacy scores of a given treatment $t_i$ for ranking all treatment compounds and identifying the effective ones, the estimated efficacy score $e_{t_i}$ is calculated as the median of all dose-dependant efficacy scores $e_{t_i}^{c_j}$ for compound $t_i$ which are greater or equal to cutoff threshold $\zeta$; in case no such $e_{t_i}^{cj}$ exists, the median of all $e_{t_i}^{cj}$ is calculated. {The compounds identified as effective against SARS-CoV-2 by DEEMD} are those for which at least one dose-dependent efficacy score is higher than $\zeta$ or equivalently:  $\mathcal{E}_{\mathcal{M}(\theta)} = \{t_i|e_{t_i} \geq \zeta\}$.

\subsection{MIL infection localization}\label{infectionlocalization}
{Localization of the infected region is achieved using the estimated $\mu_i^{p_j}$. To form the infection map $A_i$ all patches need to be aggregated. Based on the set of all patches that overlap at the pixel $x_i^{(l,m)}$, denoted as set $\mathcal{O}_{x_i}^{(l,m)}$, the value of infection map $A_{i}^{(l,m)}$ is calculated by the weighted average of its overlapping patches infection probabilities, $\{\mu_i^{p_j}\}_{{p_j} \in \mathcal{O}_{x_i}^{(l,m)}}$. We opt to set sample weights based on the sample's values, similar to the quadratic mean, but using $(\mu_i^{p_j})^{\alpha}$, where $\alpha < 1$, instead of $\mu_i^{p_j}$. The infection map $A_i$ is calculated as: 
\begin{equation}
    A_{i}^{(l,m)} =\frac{\sum_{{p_j} \in \mathcal{O}_{x_i}^{(l,m)}} (\mu_i^{p_j})^{1+\alpha}}{\sum_{{p_j} \in \mathcal{O}_{x_i}^{(l,m)}} (\mu_i^{p_j})^\alpha}.
\end{equation}}
{By using this averaging method, with $\alpha < 1$ when all {values of} $\mu_i^{p_j}$ are smaller than $1$, the average is more sensitive to higher values, and smaller values have less diminishing power on the result compared to simple averaging. This property is well-suited for infection localization since patches with high infection probabilities do not get diminished by the adjacent overlapping low probability ones. Finally, prior to rendering the infection map, a low-pass Gaussian filter, $\sigma = 60$ pixels, is applied to make the infection map smoother. For generating the infection map $A_i$ based on the MIL model $\mathcal{M}(\theta)$ we set $\alpha$ to $0.2$.}

\section{Experimental Design and Implementation}\label{section:experimental}

\subsection{Dataset}\label{sec:dataset}
\begin{figure*}[h!]
\centering
\includegraphics[width=.9\linewidth, scale=.6]{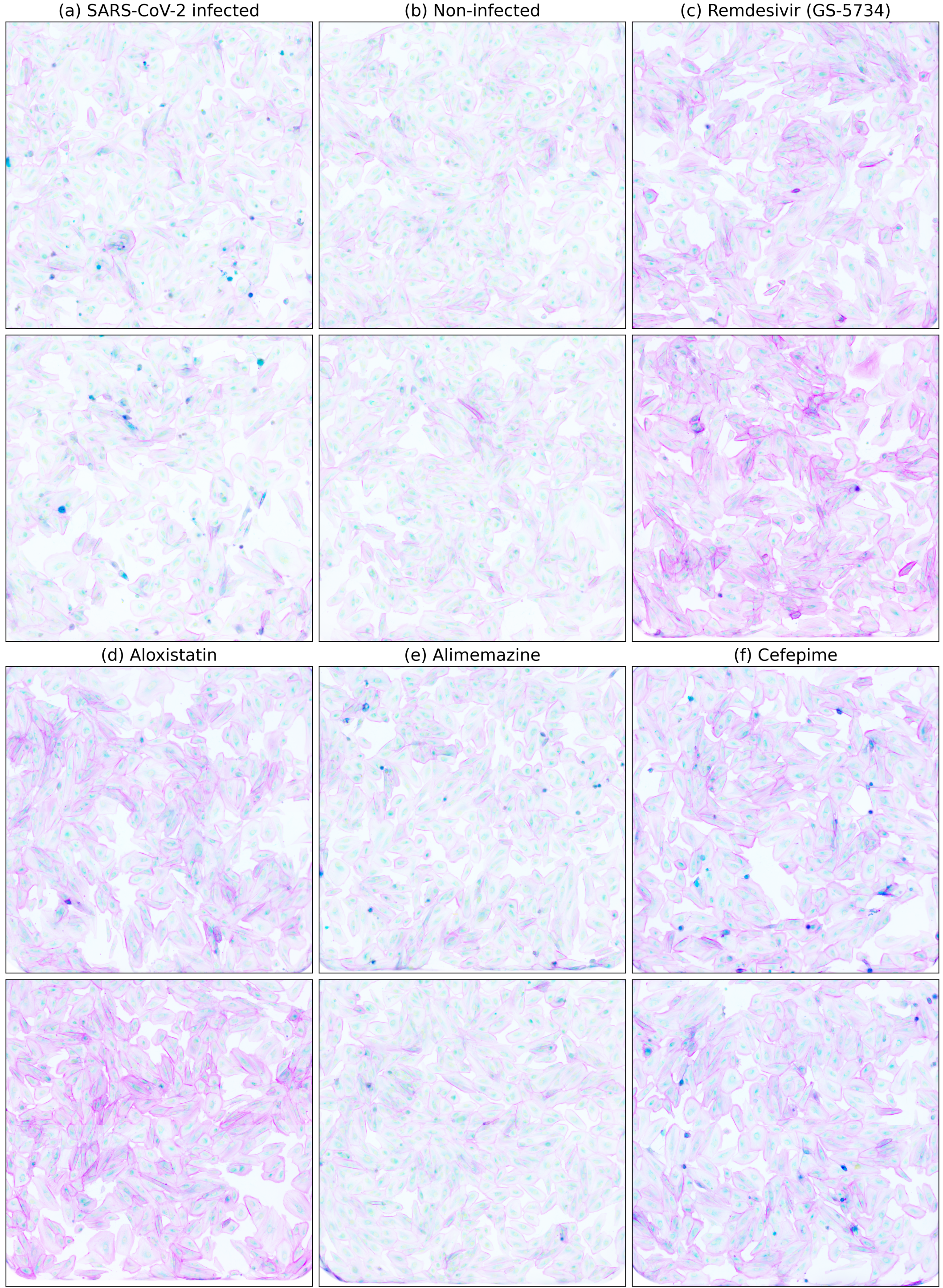}
\medskip
\vspace{-2mm}
\caption{{Example sample cell population images from the available conditions in the RxRx19a dataset: untreated SARS-CoV-2 infected (a), untreated non-infected (b), treated SARS-CoV-2 infected (c,d,e,f). Two examples are given vertically per condition. The treated examples include two treatments identified as effective by DEEMD: \textit{Remdesivir} and \textit{Aloxistatin} as well as two treatments with low efficacy scores: \textit{Alimemazine} and \textit{Cefepime}. All treatments shown are with a concentration of 0.3 $\mu$M. The red, green, and blue color channels correspond to Syto14, Phalloidin, and Hoechst stains. For better visualization, each image is color-inverted. When comparing the treated sample images with the non-infected ones for visual similarities, we note that the morphological alterations are not always observable to the human eye as discussed in Section II-B.}}\label{Fig:egcells}
\end{figure*}

We are using the publicly available RxRx19a dataset, which is the first morphological imaging dataset of cells infected with SARS-CoV-2  \cite{heiser2020identification}. The dataset consists of HRCE and African green monkey kidney epithelial (Vero) cell lines subjected to three different conditions: 1) mock non-infected control, 2) infection with ultraviolet (UV)-light inactivated SARS-CoV-2 (irradiated for 20 minutes), or 3) infection with SARS-CoV-2 at multiplicity of infection (MOI) of 0.4. All samples were incubated for 96 hours. A library of 1672 small molecules and FDA-approved treatments were applied to a subset of {samples 24 hours before being infected with} SARS-CoV-2 in 6+ half log concentrations with six replicates per dose for each compound. All cells were stained with five fluorescent dyes detecting various subcellular structures, each imaged in a separate channel on a fluorescent confocal high-content imaging microscope {(a set of representative example sample cell populations from multiple conditions and compounds can be seen in Fig~\ref{Fig:egcells})}. The dyes include Hoechst (nucleus), Syto14 (nucleoli and cytoplasmic RNA), phalloidin (actin cytoskeleton), Concanavalin A (ConA; endoplasmic reticulum) and Wheat Germ Agglutinin (WGA; Golgi and plasma membrane). The dataset consists of more than $300$K $5$-channel labeled images of size $1024 \times 1024$ and is publicly available through the  \href{https://www.rxrx.ai/rxrx19a}{\textit{Recursion} website}: \url{www.rxrx.ai/rxrx19a}. 

For this work, we only used the HRCE cell line. {For cross-fold evaluation, the untreated HRCE sample images were split into 5 folds. Each fold is further split into 3 non-overlapping sets: {1)} training, {2)} validation, and {3)} untreated test set consisting of roughly 20K, 5K, and 6K sample images respectively. All three sets are balanced in terms of class labels. The performance of the tuned model, based on the validation set, is measured on the held-out test set. The treated test set is solely used for estimating the efficacy of the treatments and remains the same across the folds.}
{We} merged the mock control and the UV inactivated control into a single non-infected class, as no cytopathic effects were observed in either condition  \cite{heiser2020identification}. {We further verified this hypothesis by using each sample image's} corresponding deep learning embeddings included in the dataset {to train a classifier to distinguish these two classes}. {To this end, we used  gradient boosted decision trees (GBDT) classifiers. GBDT is known for its effectiveness and accuracy in classification problems due to its generalized loss function \cite{friedman2001greedy}. In the same 5-fold cross-validation setting, we filtered out the SARS-CoV-$2$ infected samples. This resulted in a balanced training set of size $11$K of the mock and the UV inactivated samples. We used the \textit{scikit-learn}~\cite{scikit-learn} implementation of  GBDT with $20$ estimators, and set all other parameters to their \textit{scikit-learn} defaults. We measured the classifier's performance on the test set. The average performance of the trained classifier on the test set across the 5-folds is shown in Table~\ref{tab:UV-result}. As we can see, the classifier is not able to properly distinguish the two classes. {Thus, we accept our hypothesis about these two sample groups and merge them into one class. We further exclude sample images which did not contain any detectable cells from the HRCE samples ({see Section~\ref{cell-count}}).}}

\begin{table}[h]
\begin{center}
 \begin{tabular}{||c| c| c| c||} 
 \hline
 Condition & Precision  &  Recall & F$_1$-score\\ [0.5ex] 
 \hline\hline
  Mock & $0.46 \pm 0.25$ & $0.30 \pm 0.21$& $0.34 \pm 0.24$\\
 \hline
 UV Inactivated &$0.49 \pm 0.07$&$0.78 \pm 0.32$&$0.60 \pm 0.11$\\
 \hline
 Weighted avg & $0.48 \pm 0.15$& $0.49 \pm 0.09$& $0.42\pm 0.13$\\
 \hline
\end{tabular}
\end{center}
\caption{{Performance of the trained GBDT models on 5-fold cross-validation.}}
    \label{tab:UV-result}
\end{table}


\subsection{Baseline Classification Models}\label{Sec:Classification Baseline}
To compare the MIL model $\mathcal{M}(\theta)$ performance, two other deep learning models are trained: \textbf{1)} A model that is trained with the conventional training procedure for training convolutional neural networks by minimizing a cross-entropy-based loss function, similar to equation~(\ref{eq:loss}), with full resolution sample images as input. Denoted as  $\mathcal{W}(\phi)$, this model will be referred to as the whole-image based model $\mathcal{W}(\phi)$ throughout the paper.  Unlike most deep neural network classifiers, the input sample image is not down-sampled, and the full resolution version was used to make the comparison fair. The whole-image based model $\mathcal{W}(\phi)$ takes in the sample image $x_i$ and calculates infection probability for that cell population. \textbf{2)} Instead of limiting the model to only use $k$ instances with the highest probabilities within each bag for training, we can have a reasonable alternative approach of using all instances in the bag and assigning them with the bag-level label. There are studies in histopathology such as  \cite{coudray2018classification} that used this alternative approach and their proposed model outperformed other models at their respective tasks. We refer to this model as the patch-based model $\mathcal{V}(\psi)$ in the text. Similar to MIL instance classifier, {patch-based model} $\mathcal{V}(\psi)$ calculates a patch infection probability for all patches in the input sample image $x_i$. Intuitively, in this approach the model would potentially be able to eliminate the effect of noisy labels and estimate the true distribution of the instance labels given large enough training data. This approach is an extreme case of the MIL training procedure in which $k=N${,} where $N$ is the total number of instances within a bag.

\subsection{Implementation Details}\label{sec:implementation}
{For evaluating the performance of the models on both validation and untreated test set, we used $\eta = 0.5$. For the treatment efficacy estimation, $\zeta$ was set to $0.5$, inspired by the concept of effective concentration $50$ (EC$_{50}$) in dose-response curves, and a confidence level of $95\%$ was used in the sign test for estimating the confidence interval for the point estimated median used in the efficacy score estimations.}

{For both the MIL model $\mathcal{M}(\theta)$ and the patch-based model $\mathcal{V}(\psi)$, we used a uniform grid of $256\times256$ patches with $50\%$ overlap, resulting in $49$ patches per sample image. Since we wanted DEEMD to be applicable to other microscopy imaging datasets with minor modifications, we choose a uniform grid of patches over a cell-based segmentation map. Although a cell-based segmentation map could potentially provide a single cell resolution infection map, it is heavily dependent on the stains used in the imaging procedure, which in turn reduces the pipeline generalization. Based on the validation set, we set the hyper-parameter $k$ for the MIL model $\mathcal{M}(\theta)$ to $2$. The whole-image based model $\mathcal{W}(\phi)$ was trained on full-resolution ($1024\times1024$) sample images without any input downsampling.}   

All models are based on ResNet34 architecture, pretrained on ImageNet with a modified input layer to accommodate 5-channel inputs  \cite{he2016deep,imagenet_cvpr09}. Each model was trained for 50 epochs, monitored for early stopping, with Adam optimizer at a learning rate ${\alpha} = 10^{-4}$, {decay rates} $\beta_1 = 0.9 $ and $\beta_2 = 0.999$  \cite{kingma2014adam}, with a batch size of 128. All hyper-parameters were tuned with respect to the model's performance on the validation set. Sample images are channel normalized using the empirical means and standard deviations calculated on the training set. We used \textit{PyTorch} framework  \cite{paszke2017automatic} for model training and evaluation and treatment efficacy estimation was coded using \textit{R}  \cite{Rlang}.

{We trained the models on a single Nvidia \textit{Titan V} GPU. Note that the effective training set for the MIL model $\mathcal{M}(\theta)$ on each training iteration is ${k}/{N}$ of the whole training samples, thus its training is ${N}/{k}$ times faster compared to the patch-based model $\mathcal{V}(\psi)$. The patch-based model $\mathcal{V}(\psi)$ even has longer training time than the whole-image based model $\mathcal{W}(\phi)$ since the patches are overlapping. Thus, the MIL model $\mathcal{M}(\theta)$ is computationally more efficient compared to the other two.}

{In the early stages {of training} a deep learning model within MIL framework, the model has not learned the concepts of positive and negative instances from the training samples. Thus, the selected set $\mathcal{K}_{i}^k$, populated based on the initial exhaustive inference at the beginning of each iteration, consists of randomly selected patches from each image. To prevent this random selection process from heavily affecting the training procedure and avoiding any drastic random changes in the weights, it is important to control the learning rate of the model. To this end, we used {a} one-cycle policy based on cosine annealing \cite{smith2018disciplined} for the learning rate during the training to make sure that the network gradually learns to distinguish between the positive and negative samples. Using this {approach resulted} in a stable training procedure with consistently comparable trained models.}

\subsection{Selecting the optimal value for hyper-parameter k}\label{k-selection}
\begin{figure}[h!]
\begin{center}
\includegraphics[width=\linewidth]{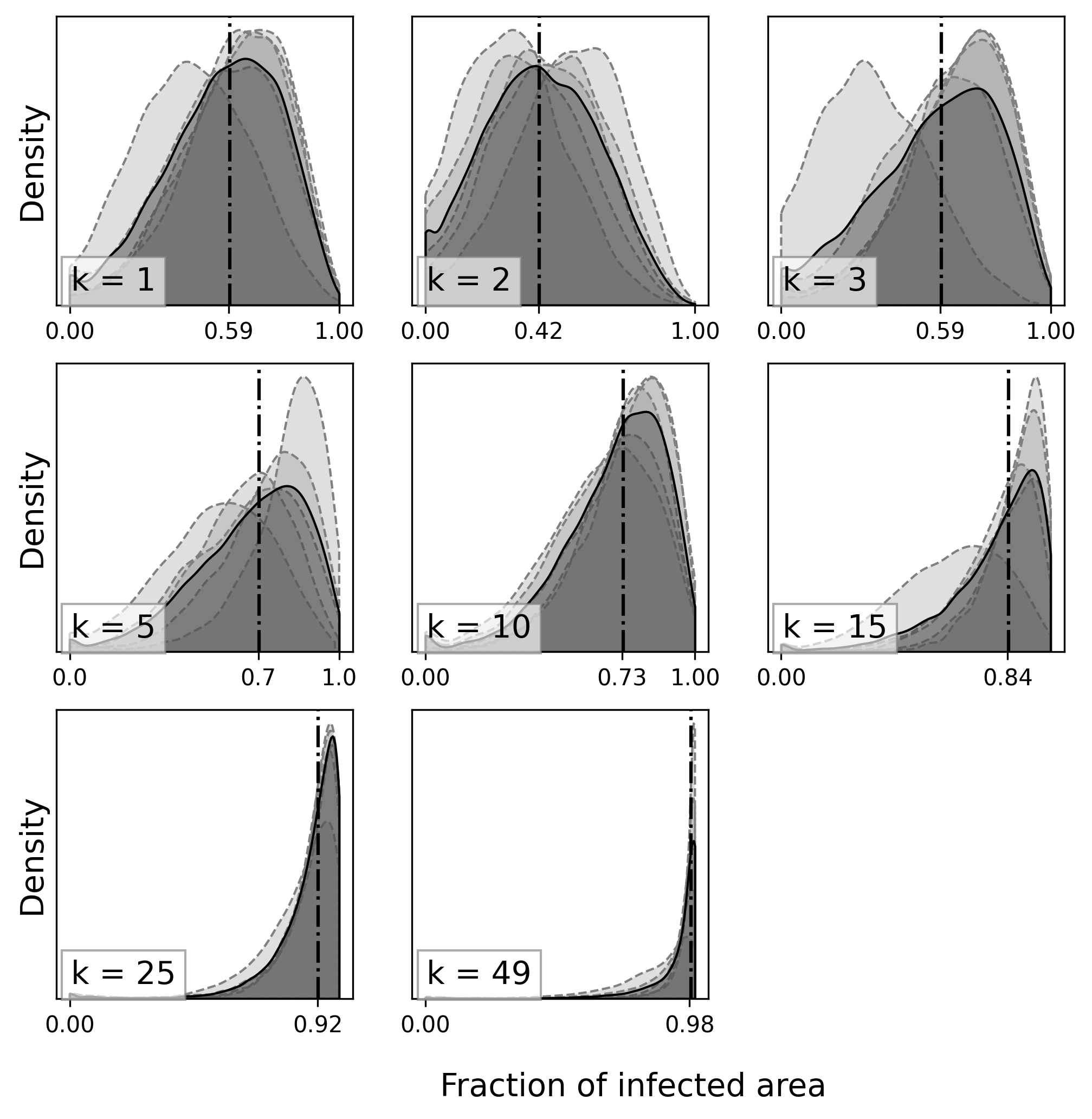}
\end{center}
\caption{{The kernel density estimation (kde) plots for distributions of the fraction of infected pixels in each sample image in the validation set (shaded plots represent each fold and the darker ones represents the aggregation over folds) for multiple values of $k$, along with the mean of the distribution.}}\label{fig:infectiondist}
\end{figure}

{In a simple yet reasonable modeling, let's assume that the virions enter the cells independent of each other and $X$ is a random variable that counts the number of virions entering a cell. Since we are dealing with a counting process, it is reasonable to assume that random variable $X$ comes from a Poisson distribution with parameter $m$: $X\sim \text{Poisson}(\text{m})$, where $m$ is the multiplicity of infection (MOI). The probability of a cell being infected is calculated as follows:
\begin{equation}
    \mathbb{P}[\text{A cell is infected}] = 1 - \mathbb{P}[X = 0]= 1 - e^{-\text{m}}.
\end{equation}
}
{Based on the details provided in \cite{heiser2020identification}, in the preparation process of the the RxRx19a dataset, the samples were infected with a MOI of $0.4$. Based on our modeling of the infection process, this value would indicate that around $33\%$ of the cells in each sample are infected on average.}
{We incorporate this domain knowledge about the expected number of infected cells within each sample image in selecting the optimal value for hyper parameter $k$. Since we are using a uniform grid of overlapping patches for the sample images, the number of the selected patches in the set $\mathcal{K}_i^k$ can not be directly translated into the fraction of the infected cells within each sample image. The area of the infected regions based on the infection maps can be a good approximation for the faction of the infected imaged cells. We trained multiple MIL models with different values for the hyper-parameter $k \in \{1,2,3,5,10,15,25,49\}$. Next, we calculated the fraction of pixels that were higher than the cutoff threshold $\eta$ in the generated infection map for each sample images in the validation set for all of the MIL models, as shown in Fig~\ref{fig:infectiondist}. We can see that when $k$ equals to $2$, the average of the distribution is close to the theoretical calculated value using the MOI, $33\%$, hence, it is the optimal value for $k$. In the rest of the paper, unless otherwise stated, DEEMD refers to the model trained with $k=2$.}

\section{Results and Discussion}\label{sec:res}

\subsection{MIL can accurately predict SARS-CoV-2 infection}\label{sec:classification}
\begin{figure}[h!]
\vspace{-15mm}
\begin{center}
\includegraphics[width=\linewidth]{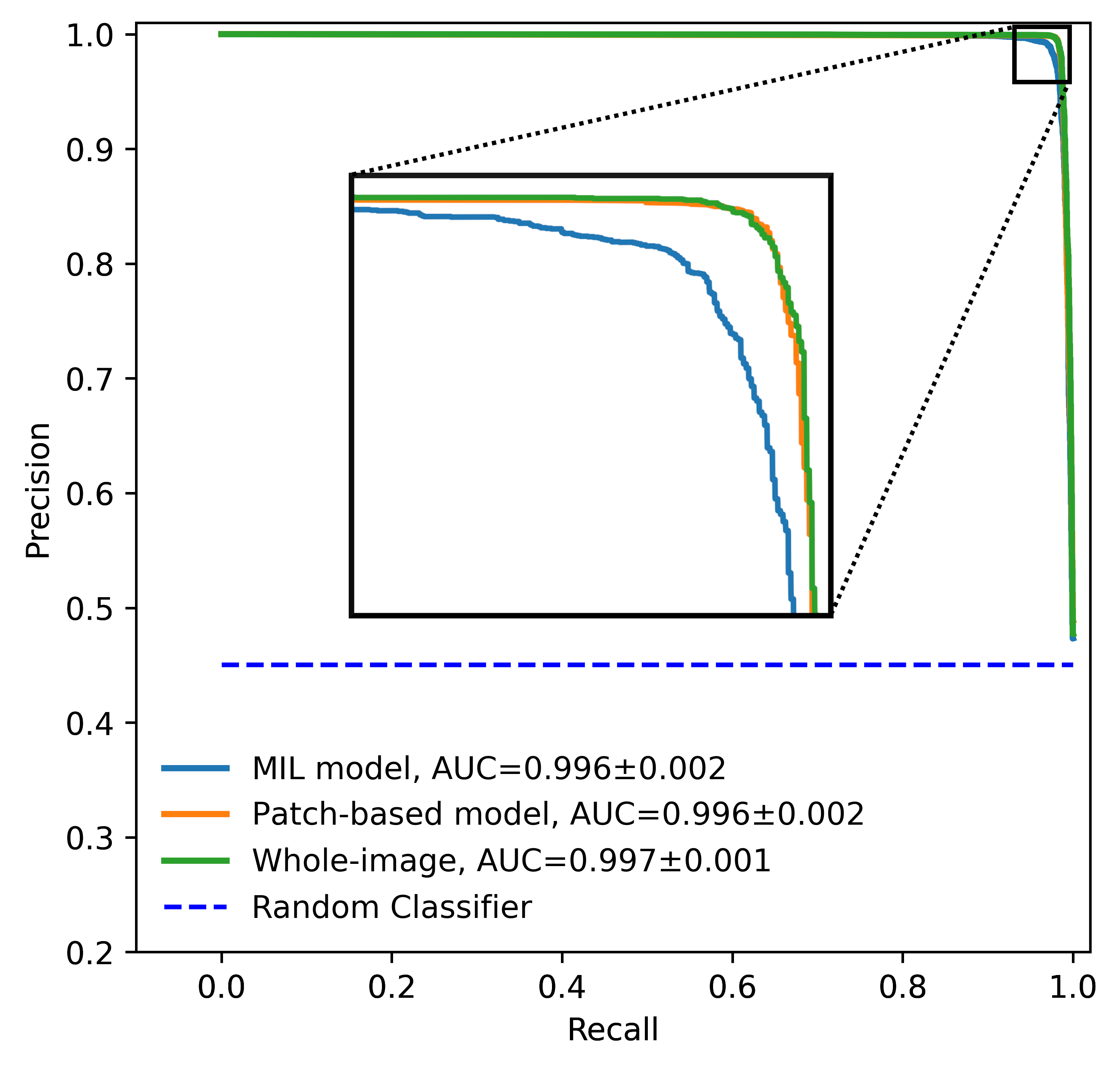}
\end{center}
\caption{Comparison of precision-recall curves. All three models are capable of accurately classifying the sample images as infected or non-infected in the untreated test set. A random classifier is included as a reference.}\label{fig:PrRcCurve}

\end{figure}

We evaluate the performance of the MIL model $\mathcal{M}(\theta)$, along with two baseline models, on the untreated test set in terms of the area under the curve for precision-recall curves, or average precision, as shown in Fig~\ref{fig:PrRcCurve}. The precision-recall curve effectively represents the trade-off between precision and recall for all possible cut-off values based on the model predictions. All three models are capable of accurately classifying the sample images into non-infected and SARS-CoV-2 infected classes, with an average precision of $\approx 0.99$. The performance of the models is independent of the learner architecture; similar results were observed with models based on VGG$16$ architecture  \cite{simonyan2014very}. High average precision implies that these models have learned morphological features that can be generalized to the untreated test set for accurate classification, however, the learned feature spaces {in each of these models} have drastically different characteristics due to their input and training procedures. The MIL model $\mathcal{M}(\theta)$ has learned to extract highly discriminative features {of} micro-populations or single cells {from patches of the original images} whereas the whole-image based model $\mathcal{W}(\phi)$ focuses on the macro-population of cells and learns morphological features of the population, which are not as detailed as in the MIL model $\mathcal{M}(\theta)$. The patch-based model $\mathcal{V}(\psi)$ lies in between those two models in the spectrum; on the one hand it has been trained using patches with micro-population features, but on the other hand, the labels it was provided for training were noisy {and it was up to the model to distinguish the noisy labels and discard them}. Thus, this model is not focused on the details as the MIL model $\mathcal{M}(\theta)$. These shifts in the learned feature spaces are further transferred to the downstream analysis of treatment efficacy, as discussed in Section \ref{sec:candidate_compounds}. {It should be noted that the performance of the models in the classification task serves as a sanity check to ensure that the trained models are properly trained, i.e. generalizable to the unseen untreated test set, and should not be considered the definitive criterion for model selection. As we demonstrate in Section~\ref{sec:candidate_compounds}, structured estimated efficacy scores of these models are the main criteria for model selection.}

\begin{table}[H]
\centering
\begin{tabular}{lllllllll}
\hline
k         & 1     & 2     & 3     & 5     & 10    & 15    & 25    & 49    \\ \hline
Pr & 0.989 & 0.971 & 0.955 & 0.905 & 0.957 & 0.893 & 0.795 & 0.481 \\
Rc    & 0.971 & 0.987 & 0.982 & 0.990 & 0.991 & 0.992 & 0.994 & 0.994 \\ \hline
\end{tabular}
\caption{{Average precision and recall values for a 5-fold cross validation across multiple choices of k, as outlined in the Section~\ref{sec:dataset}, with a threshold of $\eta = 0.5$ for inference (Eq. (3)).}}\label{tab:pr-across-k}
\end{table}

\subsection{DEEMD dose-dependent efficacy scores are well-structured}\label{sec:candidate_compounds}

\begin{figure*}[h!]
\centering
\begin{subfigure}[t]{\textwidth}
\centering
\subcaption[]{DEEMD dose-dependent efficacy scores based on MIL model $\mathcal{M}(\theta)$ with $k=2$ \hfill { }}
\vspace{-8mm}
\includegraphics[width=\linewidth]{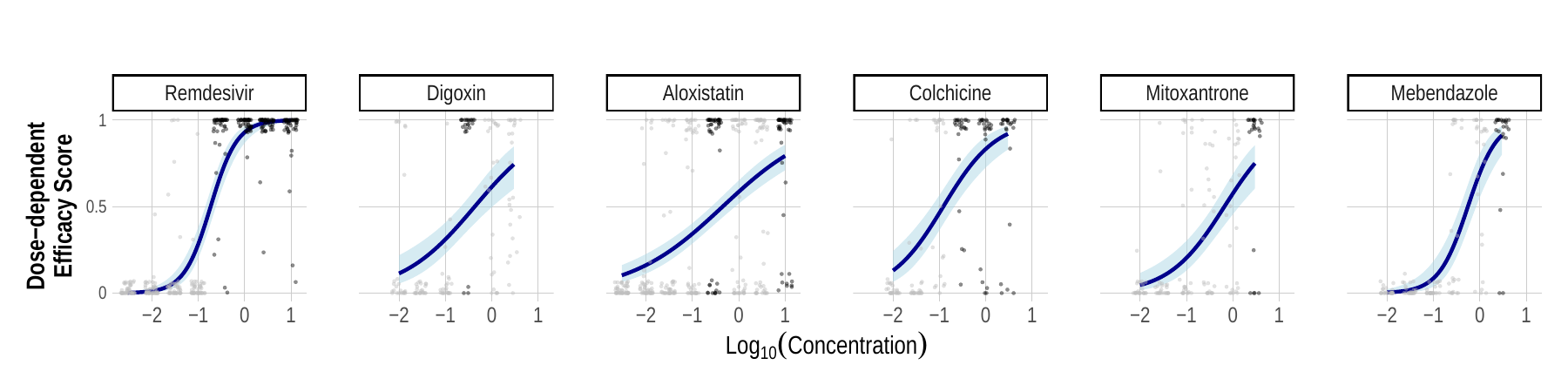}
\label{fig:dose_M}
\end{subfigure}
\\
\vspace{-5mm}
\begin{subfigure}[t]{\textwidth}
\centering
\subcaption[]{Dose-dependent efficacy scores based on the patch-based model $\mathcal{V}(\psi)$ \hfill { }}
\vspace{-8mm}
\includegraphics[width=\linewidth]{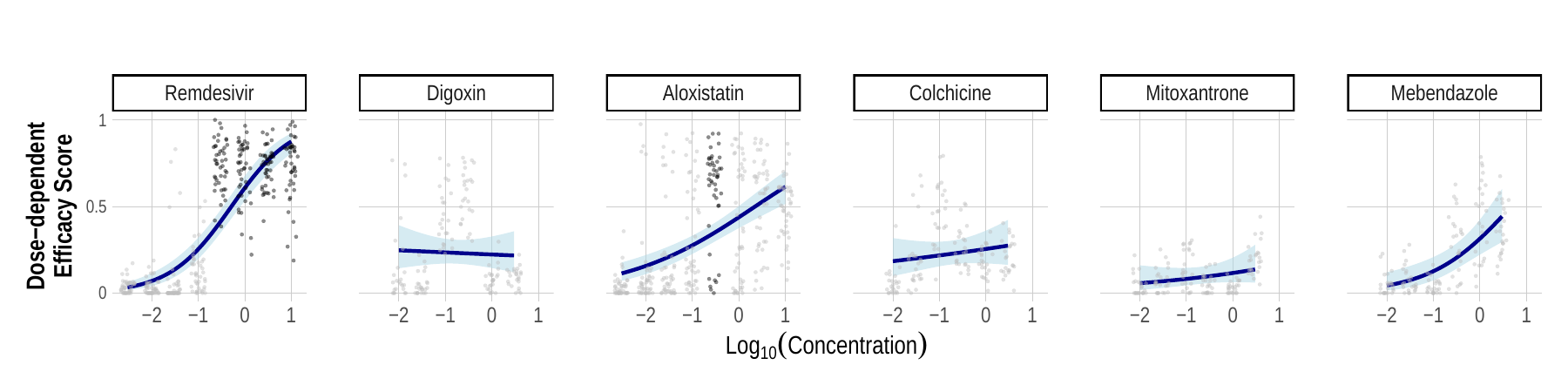}
\label{fig:dose_V}
\end{subfigure}
\\
\vspace{-5mm}
\begin{subfigure}[t]{\textwidth}
\centering
\subcaption[]{Dose-dependent efficacy scores based on the whole-image based model $\mathcal{W}(\phi)$ \hfill { }}
\vspace{-8mm}
\includegraphics[width=\linewidth]{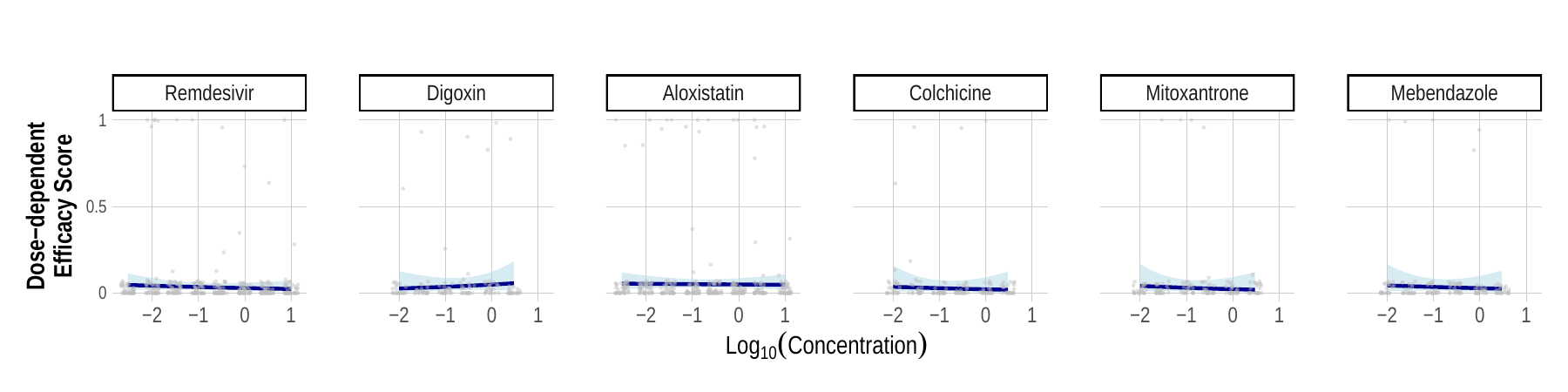}
\label{fig:dose_W}
\end{subfigure}
\begin{subfigure}[t]{\textwidth}
\centering
\subcaption[]{\hfill { }}
\vspace{-15mm}
\includegraphics[width=\linewidth]{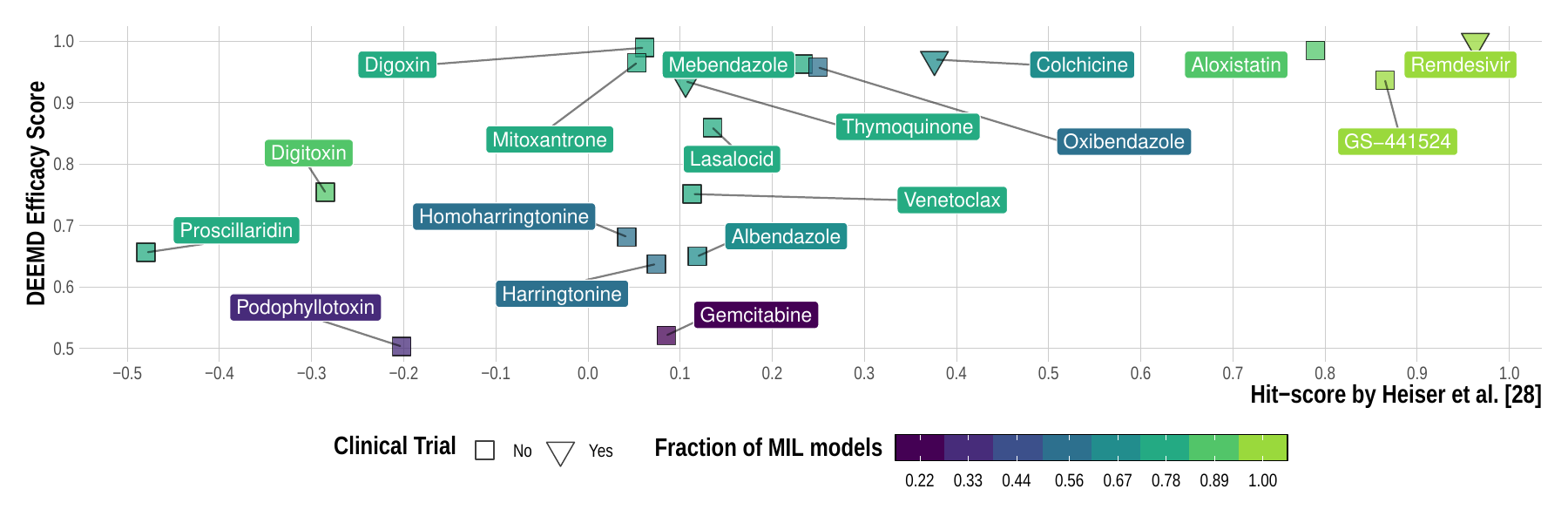}
\label{Fig:hit_plot}
\end{subfigure}
\vspace{-7mm}
\caption{DEEMD estimated efficacy scores for the identified treatments in $\mathcal{E}_{\mathcal{M}(\theta)}$. \textbf{(\subref{fig:dose_M})} Estimated Dose-dependent efficacy scores for top-ranked treatments by the MIL model $\mathcal{M}(\theta)$. The x and y axes show $\text{log}_{10}(\text{Concentration})$ and dose-dependent efficacy score for each compound. Each data point corresponds to a replicate sample for $t_i^{c_j}$. The points for a concentration are black if DEEMD identified that concentration to be effective, i.e. $e_{t_i}^{c_j} \geq \zeta$. For better visualization and to avoid overlapping points, a small noise on both x and y axis are added to the data points.  We used all of the data points, $\mathcal{T}_{t_i}$, to fit a logistic regression to better visualize the trend. {\textbf{(\subref{fig:dose_V})} and \textbf{(\subref{fig:dose_W})} are similar plots for predictions based on the whole-image based model $\mathcal{W}(\phi)$ and the patch-based model $\mathcal{V}(\psi)$ respectively.} \textbf{(\subref{Fig:hit_plot})} Comparison of DEEMD efficacy scores and Heiser \textit{et al.} hit-scores  \cite{heiser2020identification} for treatments in $\mathcal{E}_{\mathcal{M(\theta)}}$. Each data point is color-coded based on the fraction of the MIL models that identify it as effective and its shape indicates whether the treatment has gone into clinical trial against COVID-19. The y-axis represents the DEEMD efficacy scores and hit-scores are on the x-axis.}\label{Fig:panel}
\end{figure*}

We assume that an efficacious treatment is able to effectively stop viral infection and prevent major infection-induced morphological changes in the cell population. Using morphological analysis we can estimate treatment efficacy by profiling the treated and infected cell morphology and quantifying its similarity to non-infected and infected morphology.  We applied DEEMD to the infected and treated cell images from the RxRx19a dataset, using $\zeta = 0.5$ inspired by concept of EC$_{50}$ in dose-response curves, which resulted in a ranked list of 18 potential efficacious treatments against SARS-CoV-2, $\mathcal{E}_{\mathcal{M}(\theta)}$ {Table~\ref{tab:compounds}}. The dose-dependent efficacy scores for top 6 ranked effective compounds identified by DEEMD are shown in {Fig~\ref{Fig:panel}-(\subref{fig:dose_M})}.

\begin{table}[h]
\begin{center}
 \begin{tabular}{||c| c| c| c||} 
 \hline
 Rank & Compound &  Studies \\ [0.5ex] 
 \hline\hline
 1 & Remdesivir  &  \cite{yousefi2020repurposing,heiser2020identification,cuccarese2020functional,mirabelli2020morphological,tzou2020coronavirus,li2021remdesivir,pruijssers2020remdesivir,xie2020nanoluciferase,yang2020repurposing,zandi2020repurposing,choy2020remdesivir}\\ 
 \hline
 2 & Digoxin  & \cite{tzou2020coronavirus,cho2020antiviral} \\
 \hline
 3 & Aloxistatin  & \cite{ou2020characterization,yousefi2020repurposing,shang2020cell,olaleye2020discovery,heiser2020identification,cuccarese2020functional} \\
 \hline
 4 & Colchicine &  \cite{yousefi2020repurposing,CumhurCure2020,tardif2021efficacy,hariyanto2021colchicine,schlesinger2020colchicine}\\
 \hline
 5 & Mitoxantrone &  \cite{Zhang2020}\\ 
 \hline
 6 & Mebendazole &  \cite{wang2020identification,farag2020identification,law2020identifying} \\ 
 \hline
 7 & Oxibendazole &  \cite{law2020identifying}\\ 
 \hline
 8 & GS-441524 &   \cite{shi2021preclinical,li2021remdesivir,pruijssers2020remdesivir,xie2020nanoluciferase,yang2020repurposing,zandi2020repurposing}\\ 
 \hline
 9 & Thymoquinone &  \cite{ahmad2020covid,elgohary2021thymoquinone,sommer2020thymoquinone,xu2021computational}\\ 
 \hline
 10 & Lasalocid &  \cite{svenningsen2021ionophore} \\ 
 \hline
 11 & Digitoxin &  \cite{jeon2020identification,ko2021comparative,pollard2020classical}\\ 
 \hline
 12 & Venetoclax &  \cite{furstenau2020covid}\\ 
 \hline
 13 & Homoharringtonine &    \cite{yang2020repurposing,chen2020high,choy2020remdesivir,ianevski2020potential,yousefi2020repurposing,ianevski2020identification,wen2021proposal}\\ 
 \hline
 14 & Proscillaridin &    \cite{gao2020repositioning}\\ 
 \hline
 15 & Albendazole &   \cite{law2020identifying}\\ 
 \hline
 16 & Harringtonine &  \cite{hu2022harringtonine} \\ 
 \hline
 17 & Gemcitabine &   \cite{yang2020repurposing,zhang2020gemcitabine,ianevski2020identification}\\ 
 \hline
 18 & Podophyllotoxin &   \cite{hensel2020challenges}\\ 
 \hline
\end{tabular}
\end{center}
\caption{{Ranked list of DEEMD identified treatments, $\mathcal{E}_{\mathcal{M(\theta)}}$, along with drug repurposing or clinical studies that reported to show effectiveness against SARS-CoV-2 or COVID-19.}}
    \label{tab:compounds}
\vspace{-5mm}
\end{table}

As is expected for effective antiviral compounds, the dose-dependent efficacy score of the identified treatments increases with increasing concentration, similar to the fitted logistic curve. We expect to observe higher effectiveness with increasing concentration of a treatment with antiviral potency against a specific target, up to the point that it does not cause toxicity or drastically alter cell morphology. Importantly, many of the identified compounds were previously demonstrated to have antiviral activity against SARS-CoV-2 by other drug repurposing studies for COVID-19 based on morphological profiling  \cite{mirabelli2020morphological,heiser2020identification,cuccarese2020functional}, supporting DEEMD's methodology. On the other hand, {the whole-image based model $\mathcal{W}(\phi)$ completely fails to identify any compound using the same pipeline. We would expect the efficacy scores generated by any efficacy estimation model to exhibit some form of continuity when compared for the same compound across multiple increasing or decreasing concentrations, in simple 3 or 5 monotonic parameter models or more complex bi-phasic or bell-shaped ones. {Figs.~\ref{Fig:panel}-(\subref{fig:dose_M},\subref{fig:dose_V})}  demonstrate that this form of continuity could be quantitatively observed in the MIL model $\mathcal{M}(\theta)$ dose-dependent efficacy scores, whereas the scores from the whole-image model $\mathcal{W}(\phi)$ lack any form of this continuity even for the well-studied, known efficacious compounds such as Remdesivir ({Fig.~\ref{Fig:panel}-(\subref{fig:dose_W})}).} As mentioned in Section~\ref{sec:classification}, the patch-based model $\mathcal{V}(\psi)$ is a hybrid of the other two models in terms of its training. This hybridization also manifests itself in the distribution of the estimated efficacy scores, see {{Fig~\ref{Fig:panel}-(\subref{fig:dose_V})}}. We can see that the estimations are slightly structured, not as much as DEEMD's. Implying that the model was not able to extract the informative features and information completely due to its noisy training environment.

\vspace{-2mm}
\subsection{DEEMD-identified treatments are reoccurring in the literature}\label{sec:mechanism}
DEEMD has identified treatments that are reported in the literature to possess therapeutic activity against SARS-CoV-2 or COVID-19 \cite{mirabelli2020morphological,heiser2020identification,cuccarese2020functional,yousefi2020repurposing}. In {Fig~\ref{Fig:panel}-(\subref{Fig:hit_plot})} we compare DEEMD efficacy scores to hit-scores reported by Heiser \textit{et al.} \cite{heiser2020identification}, which similarly used the RxRx19a dataset. We can see that reoccurring treatments with high DEEMD efficacy scores, namely \textit{Remdesivir}, \textit{GS-441524}, and \textit{Aloxistatin} were also assigned a high hit-score by Heiser \textit{et al.}\cite{heiser2020identification} ($\rho = 0.56,p = 0.02$). In the following, we review DEEMD top ranked identified treatments and briefly discuss their potential mechanisms of action against SARS-CoV-2 infection. A complete list of treatments identified to be effective by DEEMD along with the studies that have reported them to be effective against SARS-CoV-$2$ or COVID-19 is presented in the {Table~\ref{tab:compounds}}. We also compare the DEEMD dose-dependent efficacy score for these compounds to previously reported EC$_{50}$ values as a measure of validity whenever such data is available. However it should be noted that due to differences in experimental design, such as cell line, timing of treatment and duration of infection, EC$_{50}$ values from cell-based studies can vary widely.

The most well-known treatment in $\mathcal{E}_{\mathcal{M}(\theta)}$ is \textit{Remdesivir} and its metabolite, \textit{GS-441524}. Previously studied for their antiviral effectiveness against Ebola virus, these compounds target the virus-encoded RNA-dependent RNA polymerase complex needed by RNA viruses to replicate their genome  \cite{gordon2020antiviral}. Multiple studies and clinical trials have found that both compounds are effective against SARS-CoV-2, which led to its emergency approval by the FDA  \cite{yousefi2020repurposing,heiser2020identification,cuccarese2020functional,mirabelli2020morphological} (and refrences within). DEEMD dose-dependent efficacy scores for \textit{Remdesivir} are shown in {Fig~\ref{Fig:panel}-(\subref{fig:dose_M})}. We can see that the DEEMD efficacy score for \textit{Remdesivir} is persistently close to $1$ for concentrations of $0.3-10$ $\mu$M which is consistent with EC$_{50}$ values reported by a number of other cell-based studies ranging between $0.003$ to $27$ $\mu$M  \cite{tzou2020coronavirus}. Similarly, \textit{GS-441524} was estimated to be efficacious at concentrations of $3-10$ $\mu$M by DEEMD, which is comparable to previously reported EC$_{50}$ values ranging from $0.5$ to $8.2$ $\mu$M  \cite{tzou2020coronavirus}.

\textit{Digoxin} is a treatment used for heart disease with a well-established safety profile. Multiple \textit{in vitro} drug repurposing studies for COVID-19 reported its ability to inhibit SARS-CoV-2 infection  \cite{tzou2020coronavirus}. The exact mechanism of action for viral inhibition is not identified yet, however, Cho \textit{et al.} \cite{cho2020antiviral} hypothesised that \textit{Digoxin} inhibition occurs at the step of viral RNA synthesis. They used multiple FDA-approved treatments, including \textit{Digoxin}, on SARS-CoV-2 infected Vero cells. To understand how the drugs might inhibit SARS-CoV-2, they were administered at three different time points: 1) prior to infection (prophylactic), 2) at the time of infection (entry), and 3) after the infection (therapeutic). They reported that \textit{Digoxin} showed high efficacy following prophylactic and therapeutic administration but failed to effectively inhibit the virus when administered at the time of infection. However, it should be noted that \textit{Digoxin} exhibited cytotoxicity, as reported by Mirabelli \textit{et al} \cite{mirabelli2020morphological}. {Fig~\ref{Fig:panel}-(\subref{fig:dose_M})} shows the DEEMD estimated dose-dependent efficacy scores for \textit{Digoxin}; it passed the threshold at $0.3$ $\mu$M which is aligned with multiple studies that found an EC$_{50}$ between $0.04$ and $0.2$ $\mu$M \cite{tzou2020coronavirus}. We hypothesize that the DEEMD dose-dependent efficacy scores are not conclusive for higher concentrations because of \textit{Digoxin} reported cytotoxicity.

\textit{Aloxistatin} (\textit{E-$64$d}) is another potential candidate treatment reported in the literature as an effective agent against SARS-CoV-2. \textit{Aloxistatin} is a membrane-permeable irreversible cysteine-protease inhibitor of calpains and cathepsins. Recent studies have shown that SARS-CoV-2 requires cathepsin L to enter some cell types. \textit{Aloxistatin} can significantly reduce entry of SARS-CoV-2 pseudovirions by inhibiting cathepsin L  \cite{ou2020characterization,yousefi2020repurposing,shang2020cell} (and references within). {Fig~\ref{Fig:panel}-(\subref{fig:dose_M})} shows DEEMD estimated dose-dependent efficacy scores for \textit{Aloxistatin}. As we can see, the estimated efficacy is higher than the cutoff threshold $\zeta$ at $0.3$ and $10$ $\mu$M. Olaleye \textit{et al.} used Vero cells to investigate the antiviral activity of several compounds including \textit{Aloxistatin} for which they reported an EC$_{50}$ of $22$ $\mu$M  \cite{olaleye2020discovery}. Two additional studies  \cite{heiser2020identification} and  \cite{cuccarese2020functional} using morphology-based approaches found that \textit{Aloxistatin} shows strong efficacy without inducing morphological changes to cells. 

DEEMD also identified \textit{Mitoxantrone} to be effective against SARS-CoV-2, as shown in {Fig~\ref{Fig:panel}-(\subref{fig:dose_M})}. The cell surface heparan sulfate (HS) is a molecule commonly found on the membrane and on extracellular proteins of cells that assists the endocytosis of many cargos, including SARS-CoV-2 spike. Recently, Zhang \textit{et al.} reported that HS facilitates spike-dependent viral entry of SARS-CoV-2 \cite{Zhang2020}. They experimented with inhibitor drugs that target the HS-dependent cell entry pathway, and observed that \textit{Mitoxantrone} inhibited viral entry by directly binding to cell surface HS. DEEMD identified \textit{Mitoxantrone} to be effective at $3.0$ $\mu$M, while  Zhang \textit{et al.} reported an EC$_{50}$ value of $0.03$ $\mu$M.

As shown in {Fig~\ref{Fig:panel}-(\subref{fig:dose_M})}, DEEMD also identified \textit{Colchicine} as an effective compound against SARS-CoV-2 at concentrations of $0.3 - 3.0$ $\mu$M. \textit{Colchicine} is an anti-inflammatory compound used for multiple indications including gout and heart disease. The anti-inflammatory activity is predicted to result from the inhibitory effect on tubulin polymerization and microtubule assembly  \cite{yousefi2020repurposing}. Due to the anti-inflammatory properties, and the known link between inflammation and severe COVID-19 outcomes, \textit{Colchicine} has been evaluated as a potential COVID-19 therapeutic with conflicting results  \cite{CumhurCure2020}. Encouragingly, a  meta-analysis of multiple clinical trials suggests \textit{Colchicine} is associated with improved outcomes for individuals with COVID-19  \cite{hariyanto2021colchicine}. To date, a direct antiviral effect of \textit{Colchicine} on SARS-CoV-2 has not been demonstrated, but tubulin has been implicated in coronavirus entry  \cite{schlesinger2020colchicine}. Our findings suggest that the antiviral activity of \textit{Colchicine} against SARS-CoV-2 warrants further investigation.

DEEMD also identified three compounds, \textit{Mebendazole}, \textit{Oxibendazole} and \textit{Albendazole} with effective concentrations ranging from $1.0 - 3.0$ $\mu$M {Fig~\ref{Fig:panel}-(\subref{fig:dose_M})}. These compounds belong to a large chemical family of \textit{Benzimidazoles} that are used to treat nematode and trematode infections  \cite{HARRIS20181373}. These three compounds specifically have not yet been demonstrated to exhibit therapeutic activity against SARS-CoV-2 or COVID-19. However, antiviral activity against other viruses has been shown using other \textit{Benzimidazole} derivatives with EC$_{50}$ values ranging from $0.02 - 90.0$ $\mu$M  \cite{tonelli2010antiviral}. Thus based on our analysis, these should be further examined for activity and possible mechanism of action against SARS-CoV-2.
\subsection{DEEMD identified treatments are robust across multiple selections of $k$}\label{sec:k_selection}
\begin{figure}[h!]
\centering
\includegraphics[width=\linewidth]{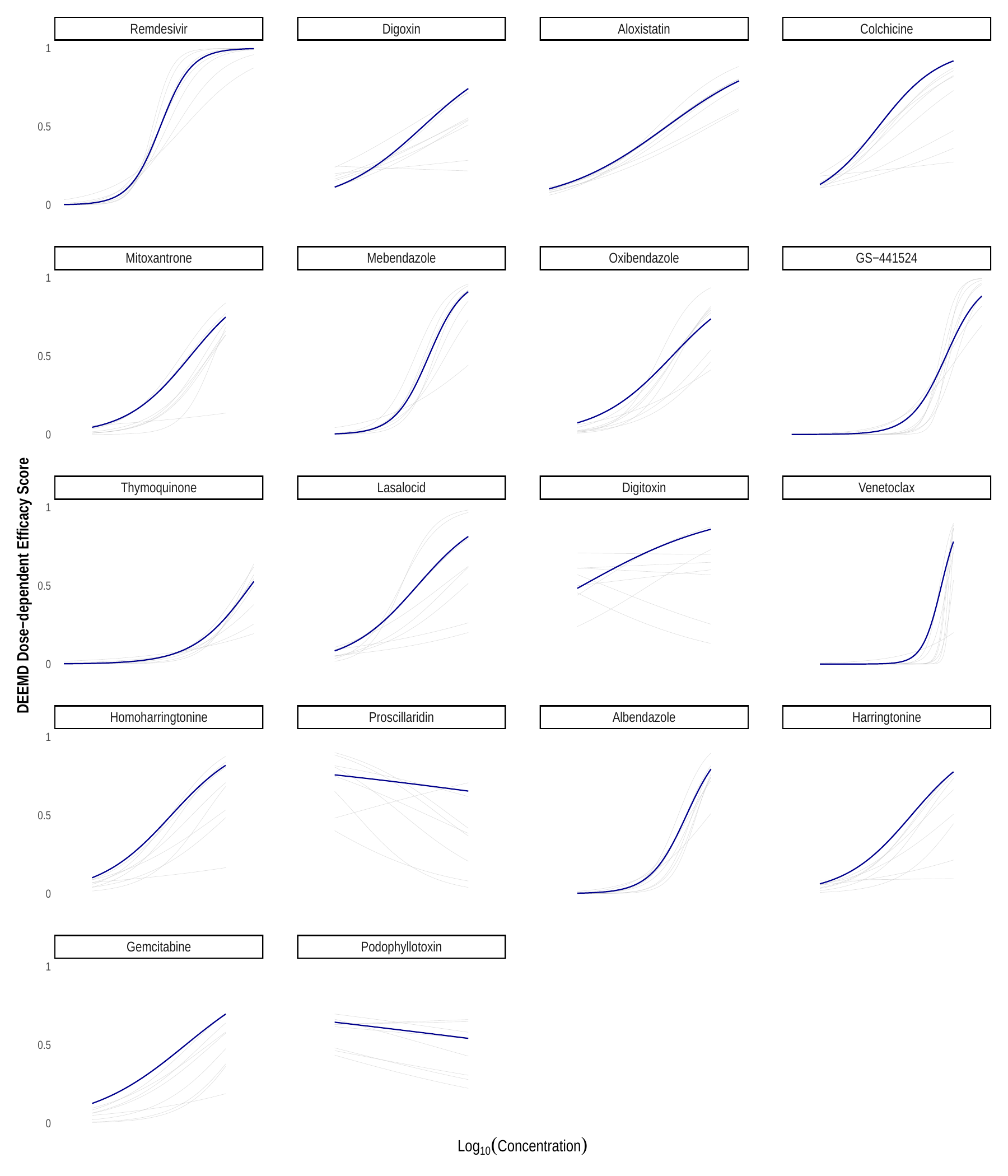}
\vspace{-2mm}
\caption{DEEMD estimated efficacy scores for identified treatments for different $k$ values. The logistic curves are fitted based on each value for $k$. The blue curves represents the MIL model with $k=2$ which showed the best classification performance on the validation set. }\label{Fig:bulk}
\end{figure}

When comparing the set of identified treatments $\mathcal{E}_{\mathcal{M(\theta)}}$ for different values of $k$, multiple reoccurring treatments were observed. Moreover, the fitted logistic regression {curve} on the estimated dose-dependent efficacy scores retained its shape and the inflection points are close together; suggesting a consensus, almost independent of choice of $k$, between the MIL models on the degree of effectiveness for various treatments (See {Section~\ref{appendix} Fig~\ref{Fig:bulk}}). As shown in {Fig~\ref{Fig:panel}-(\subref{Fig:hit_plot})}, most of the treatments in $\mathcal{E}_{\mathcal{M}(\theta)}$ have a high recurrence rate in the MIL models, implying their distinguishable ability to suppress viral-induced morphological changes robust across multiple MIL models trained with different values for $k$. We noticed that based on the patch-based model $\mathcal{V}(\psi)$ only three treatments were identified to be effective: \textit{Remdesivir}, \textit{GS-441524}, and \textit{Aloxistatin}, which are among the most commonly identified treatments both in COVID-19 drug repurposing studies and multiple MIL models trained with different $k$, as shown in {Fig~\ref{Fig:panel}-(\subref{Fig:hit_plot})}. This observation suggests that for high values of $k$ the amount of noisy labels drastically changes the dynamics of the training. The learned feature space lacks the required sensitivity for downstream analysis of treatment identification, while the model is still capable of accurately classifying the sample images into infected and non-infected classes.


\begin{figure*}[h]
\centering
\centering
\includegraphics[width=\linewidth]{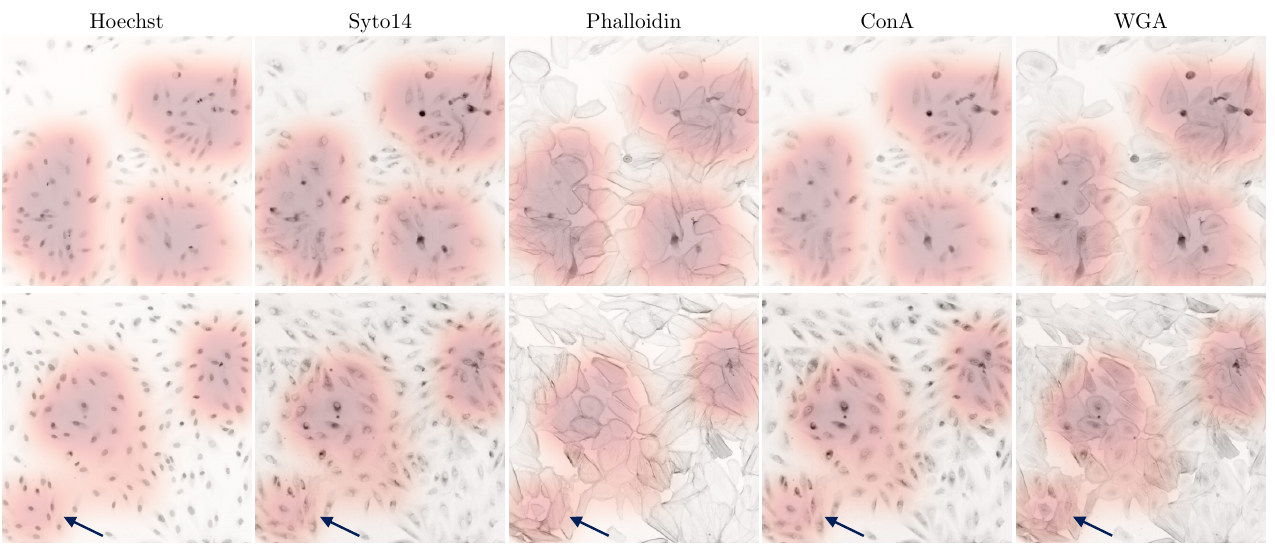}
\vspace{-5mm}
\caption{Generated infection maps for a representative set of  infected samples from the untreated test set. Sample images are overlaid with their corresponding infection map on each of the indicated stains, displayed separately. Each row represents a SARS-CoV-2 infected sample image from the untreated test set and each column corresponds to stains used in the dataset. The intensity of the red color shows the probability of the infection on each position based on the MIL model $\mathcal{M}(\theta)$ predictions. On the second row the arrows point to the same region, which is assigned a high infection probability by the MIL model $\mathcal{M}(\theta)$, despite it lacks visible signs of CPE compared to other high infection probability regions.
For better visualization, each dye image is color-inverted.}
    \label{table:InfectionSamples}
\end{figure*}

\subsection{The MIL infection maps and cytopathic effects}\label{sec:CPE}
Localization of infected regions in a sample image can be done by aggregating all estimated patch-level {infection probability predictions} $\mu_i^{p_j}$ to generate an infection map, as described in {Section~\ref{infectionlocalization}}. These maps can be used to better understand and explain the MIL model $\mathcal{M}(\theta)$ classification predictions as well as providing an annotation for patches that contain SARS-CoV-2 infected cells. Notice that in the calculation of these maps, no form of annotation was used; instead, the MIL model uses sample-level labels to localize the infection. Figure~\ref{table:InfectionSamples} shows two representative examples from the untreated test set, stain-separated overlaid with infection maps predicted by the MIL model $\mathcal{M}(\theta)$. Visual inspection of the infection maps and regions that are predicted with high infection probability, annotated as red, suggests that the patches that trigger the MIL model $\mathcal{M}(\theta)$ include but not limited to high intensity stained areas, which we predict to reflect cell death or cytopathic effects (CPE), detectable in all channels, except for Hoechst. However, there are patches without the visible signs of CPE that are assigned a high infection probability by the model, e.g. notice the identified infected region on the bottom left corner of samples in the the second row in Fig~\ref{table:InfectionSamples}. CPE refers to changes in host cell structure as a result of viral infection and SARS-CoV-2 is known to be a cytopathogenic agent  \cite{zhu2020morphogenesis}; for example SARS-CoV-2 is known to cause cell death and to induce syncytia formation (fusion of adjacent cell membranes) \cite{buchrieser2020syncytia}. CPE can be measured indirectly by using luminescent cell viability assays  \cite{riva2020large}, however, the RxRx19a dataset does not include a specific viability marker, hence CPE can not be quantified on these samples. {Nonetheless, it appears that CPE is being detected in the highlighted regions of the infection maps. These regions often contain very brightly stained rounded cells or nuclei, and we hypothesize that the stronger fluorescent signals are associated with morphological changes related to cell death such as loss of membrane integrity, cell shrinkage, and nuclear fragmentation \cite{cummings2004measurement}.} We also observed that the cell nucleus count is significantly lower in the infected samples, {Fig~\ref{Fig:cellCount} in the Section~\ref{appendix}}, which is also indicative of cell death. This suggests that the MIL model $\mathcal{M}(\theta)$ has incorporated biologically relevant morphological features into the infection map, as is expected.

\subsection{DEEMD is limited by drug toxicity}\label{sec:toxicity}
Drug toxicity refers to a compound's negative side effects on a living cell, as the compound can disrupt crucial cellular functions and pathways to the extent of causing cell death. Taking drug toxicity into consideration is essential for optimizing the concentration of the compound needed for optimal efficacy  \cite{nicholson2002metabonomics}. It is noteworthy to clarify that a low efficacy score $e_{t_i}^{c_j}$ estimated by DEEMD does not necessarily indicate dose or treatment ineffectiveness against SARS-CoV-2. If the treatment has toxic effects or induces other changes in cell morphology, the sample morphology may no longer resemble either the uninfected or SARS-CoV-2 infected class; subsequently, the model's prediction would not be conclusive. Currently, DEEMD only relies on the assumption that if the drug is toxic, the treated cell morphology would not be similar to uninfected cells, thus the model estimates a low efficacy for a toxic compound. A more complex model capable of integrating drug toxicity with cell morphology is required to properly capture the dynamics of the treatment compound, toxicity and effects on cellular morphology, including cell death.

The RxRx19a public dataset does not include images from healthy drug treated cells, which restricts the morphological feature space learned by the model during training. The model can only learn the regions corresponding to the uninfected and SARS-CoV-2 infected cell morphology, and it is unaware of the space structure outside of these regions. Including additional classes into the training dataset would allow the model to learn a wider range of morphological variations and enable it to differentiate between treatments that are ineffective from those that are impacting cell morphology through toxicity or any other mechanisms; thus a better control over the false negative rate would be in place. We noticed clear cases of drug toxicity with multiple compounds in the dataset, where high concentrations clearly disrupted cell morphology. These sample images lack a detectable signal for the different cell structures and thus we decided to exclude them from the treated test set. Training the model to identify cellular morphology associated with drug toxicity would support the identification of the compound and corresponding concentrations with optimal efficacy and minimal toxicity  \cite{nicholson2002metabonomics}.

\section{Conclusion}\label{sec:conclusion}

In this work, we present DEEMD: a pipeline capable of estimating the treatment efficacy of compounds based on morphological analysis of fluorescent-labelled cells. It includes a deep learning model trained within a MIL framework to extract morphological features corresponding to the predicted SARS-CoV-2 infection versus no infection in micro-populations, as well as generating an infection map in a weakly supervised fashion. We compared the performance of the MIL model $\mathcal{M}(\theta)$ to the conventionally trained whole-image based model $\mathcal{W}(\phi)$. As discussed in Section~\ref{sec:res}, both models are capable of accurately distinguishing between images from uninfected and SARS-CoV-2 infected sample images. By integrating a statistical test into the pipeline, DEEMD identifies efficacious compounds that have been reported to have antiviral effectiveness against SARS-CoV-2 using other methods, supporting the performance of the proposed pipeline, whereas the whole-image based model $\mathcal{W}(\phi)$ fails to estimate meaningful predictions.



In the future, we plan to apply DEEMD to more comprehensive datasets {with various disease models and cell lines} that include drug-treated, uninfected cells along with specific markers of viral infection that can be used to properly address the shortcomings and limitations of the current version of DEEMD. {Having an image channel designated to the viral-specific markers would help us verify the generated infection maps. The viral-infection marker maps would provide us with a way to quantitatively evaluate the generated infection maps relative to a positive control and investigate their relations. {It is worth mentioning that while viral-specific markers are the gold standard for localizing infections, they come with a trade-off between precision and recall. However, DEEMD is capable of localizing and generating infection maps without requiring such markers.} Having compound treated, uninfected samples would allow us to first quantify any morphology changes induced by treatment alone, which could be used to disentangle the virus- and drug-induced morphological effects on cells. Treated, uninfected samples may also be advantageous for understanding combination therapies if such data is available in the future. Exploring the compound-induced morphological changes to cells is a potential extension of DEEMD. One approach to such modeling, considering the number of possible outcomes of the classification is very large, is training a model capable of predicting the applied compound given a sample image in a self-supervised setting \cite{zbontar2021barlow,chen2020simple}. DEEMD is capable of reporting quantifiable changes based on morphological reasoning of deep neural networks. Already DEEMD has practical value to cut costs associated with drug repurposing, as our proposed pipeline can rapidly identify candidate antiviral treatments and predict efficacy against SARS-CoV-2 infection. Alongside an established supply chain and pre-existing clinical safety data, DEEMD can accelerate the drug repurposing screening process, bringing machine learning-based drug repurposing one step closer to being widely applied to therapeutics.}
\vspace{-2mm}
\section{Acknowledgments}
{The authors would like to thank the reviewers for their valuable feedback that improved the paper. The authors are grateful to the NVIDIA Corporation for donating GPUs used in this research.}  
\vspace{-3mm}

\section{Appendix}\label{appendix}

\subsection{Noisy label analysis in MIL}\label{supp:noise}

\begin{figure}[h!]
\begin{center}
\vspace{-3mm}
\includegraphics[width=1\linewidth]{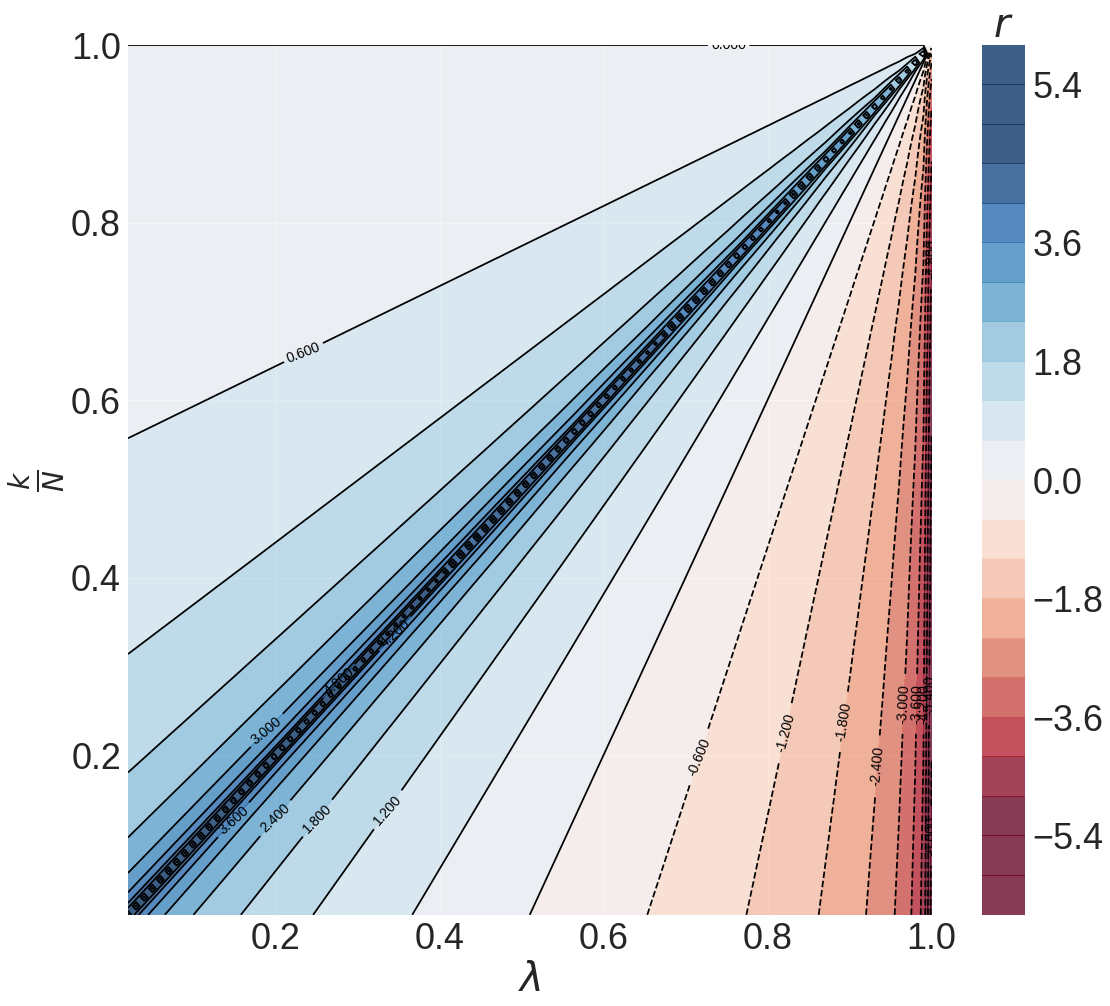}
\caption{Contour plot for $r(\lambda,\frac{k}{N})$, all possible values for $\lambda$ and $\frac{k}{N}$ are shown on the x and y axis respectively.}\label{fig:ContourPlot}
\end{center}
\end{figure}

{As mentioned in Section~{\ref{Sec:Classification Baseline}}, the patch-based model $\mathcal{V}(\psi)$ trains the model in the presence of noisy labels. To better understand this model and compare it to the MIL model $\mathcal{M}(\theta)$, it would be insightful to measure how much noise is in the training labels for both models. Each image is split into $N$ patches and there are $N_p$ positive samples and $N_n$ negative samples in the dataset. We define the label noise in the dataset to be the ratio of the samples correctly labeled in the dataset to the total number of samples. Without loss of generality, assume that the fraction of the patches in a sample image that are truly infected has an expected value of $\lambda$.
Therefore, the expected noise ratio (NR) for the MIL model $\mathcal{M}(\theta)$ and the patch-based model $\mathcal{V}(\psi)$ can be expressed as follows: 
\begin{align}
    \text{NR}_{\mathcal{M}(\theta)}= \frac{|\frac{k}{N} - \lambda|N_P}{N_P + N_N} , \quad \text{NR}_{\mathcal{V}(\psi)}= \frac{(1-\lambda)N_P}{N_P + N_N}.
\end{align}
By defining $r(\lambda , \frac{k}{N})$ to be the log ratio of $\text{NR}_{\mathcal{V}(\psi)}$ to $\text{NR}_{\mathcal{M}(\theta)}$, we can quantitatively analyze these two models behaviour in different configurations for $\lambda$ and $k$:
\begin{align}
    r(\lambda , \frac{k}{N}) = \text{ln}\Big(\frac{\text{NR}_{\mathcal{V}(\psi)}}{\text{NR}_{\mathcal{M}(\theta)}}\Big) = \text{ln}\Big(\frac{1-\lambda}{|\frac{k}{N} - \lambda|}\Big).
\end{align}
When $r(\lambda , \frac{k}{N}) \geq 0$ the MIL model $\mathcal{M}(\theta)$ has less noisy labels compared to the patch-based model $\mathcal{V}(\psi)$. The landscape of $r(\lambda , \frac{k}{N})$ is visualized in Fig~\ref{fig:ContourPlot} for closer inspection. To ensure the numerical stability of $r(\lambda , \frac{k}{N})$ and keep it bounded on the $y=x$ line, a small value $\epsilon$ was added to $\text{NR}_{\mathcal{M}(\theta)}$.}

{One can notice that in almost three forth of the cases, $r(\lambda , \frac{k}{N}) \geq 0$, meaning that the MIL model $\mathcal{M}(\theta)$ has less noisy labels in its training procedure on average, hence, the model converges faster to an optimum point. Moreover, the MIL model $\mathcal{M}(\theta)$ training procedure results in a more efficient training both in terms of computation footprint and extracted features. The model is using only ${k}/{N}$ of the training set for updating the weights. This is ${N}/{k}$ times faster than using the whole dataset in the patch-based model $\mathcal{V}(\psi)$. More importantly, the patches that are selected for training the MIL model $\mathcal{M}(\theta)$, would contain more discriminative features since they were top-ranked among all patches in their samples for their informativeness. Finally, the MIL model $\mathcal{M}(\theta)$ has the capability of incorporating domain knowledge about the problem into the training by the choice of hyper-parameter $k$. By choosing $k$ based on prior or domain knowledge, the MIL model $\mathcal{M}(\theta)$ can iteratively refine its training dataset towards for less noisy labels.}
\subsection{Cell nucleus count in the sample images}\label{cell-count}

\begin{figure}[h!]
\centering
\begin{subfigure}[t]{\linewidth}
\centering
\subcaption[]{ \hfill { }}
\vspace{-3mm}
\includegraphics[width=.8\linewidth, scale=.6]{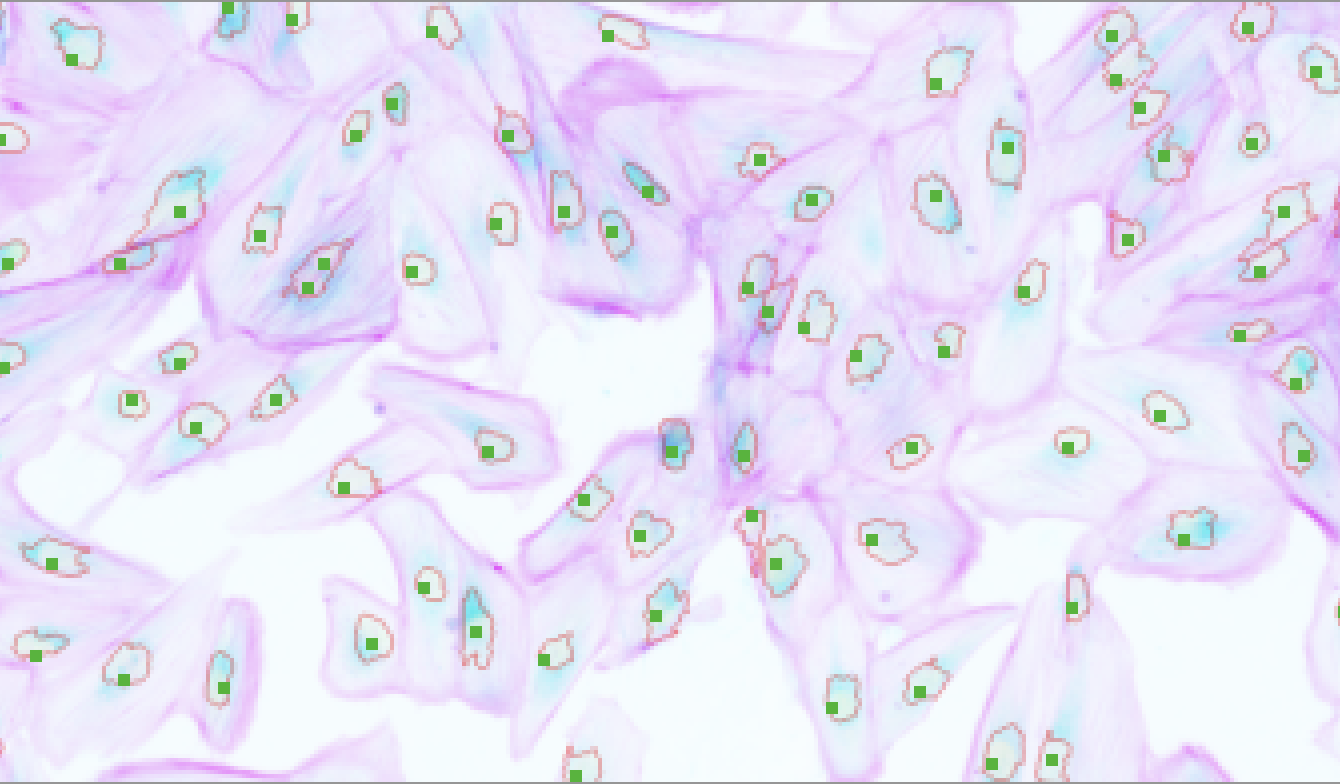}
\label{Fig:wellAnnotated}
\end{subfigure}
\\
\begin{subfigure}[t]{\linewidth}
\centering
\subcaption[]{\hfill { }}
\vspace{-5mm}
\includegraphics[width=.9\linewidth]{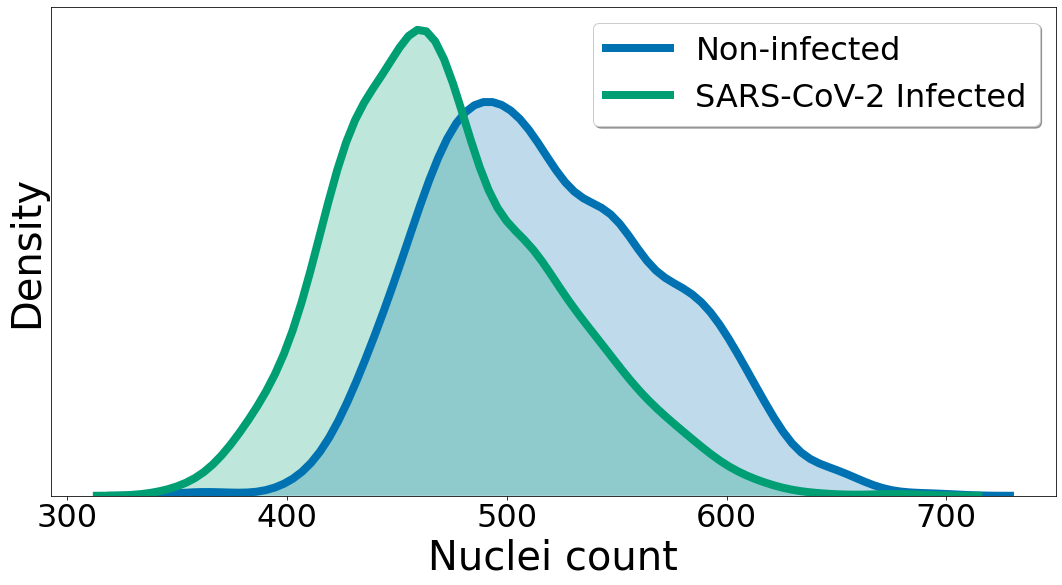}
\label{Fig:cellCount}
\end{subfigure}
\caption{{(\textbf{a}) Example annotated image of a cell population cropped for clarity. Includes the prediction and manual cell localizations as cyan outlines and pink squares respectively. The red, green, and blue color channels correspond to Syto14, Phalloidin, and Hoechst stains. For better visualization, each image is color-inverted. (\textbf{b}) Distribution of cell nucleus count in the stitched well sample images from non-infected and SARS-CoV-2 infected HRCE cells. The non-infected class consists of samples from both Mock and UV Inactivated SARS-CoV-2 classes.}}\label{Fig:cellcountpanel}
\vspace{-8mm}
\end{figure}

{The RxRx19a dataset includes Hoechst stain which binds double stranded DNA in the nucleus. We used this stain to identify and count the cells in each sample image for both explanatory analysis and preprocessing the data. We first stitch images of 4 adjacent sites of each well on each plate to reconstruct {an image of the} whole sample cell population. This step is necessary to avoid repeat counting of nuclei in multiple images. We localize and count the nuclei in each stitched image using a segmenting pipeline based on Otsu thresholding and watershed algorithm \cite{vincent1991watersheds}. {The counting pipeline performance was evaluated against 20 randomly selected manually counted stitched images with a mean absolute error of $16.0$. An example annotated image with prediction overlay is shown in {Fig~\ref{Fig:cellcountpanel}-(\subref{Fig:wellAnnotated}})}. We used the pipeline to exclude any sample images which did not contain any detectable cells from the dataset. The distribution of the nuclei count in the sample images for infected and non-infected samples {is} shown in {Fig~\ref{Fig:cellcountpanel}-(\subref{Fig:cellCount})}. The cell nucleus
count is significantly lower in the infected samples (using two-sided Mann-Whitney test with $p=0.0$).}

\subsection{Saliency maps for the whole-image model $\mathcal{W}(\phi)$}
\begin{figure*}[ht!]
\centering
\includegraphics[width=.9\linewidth]{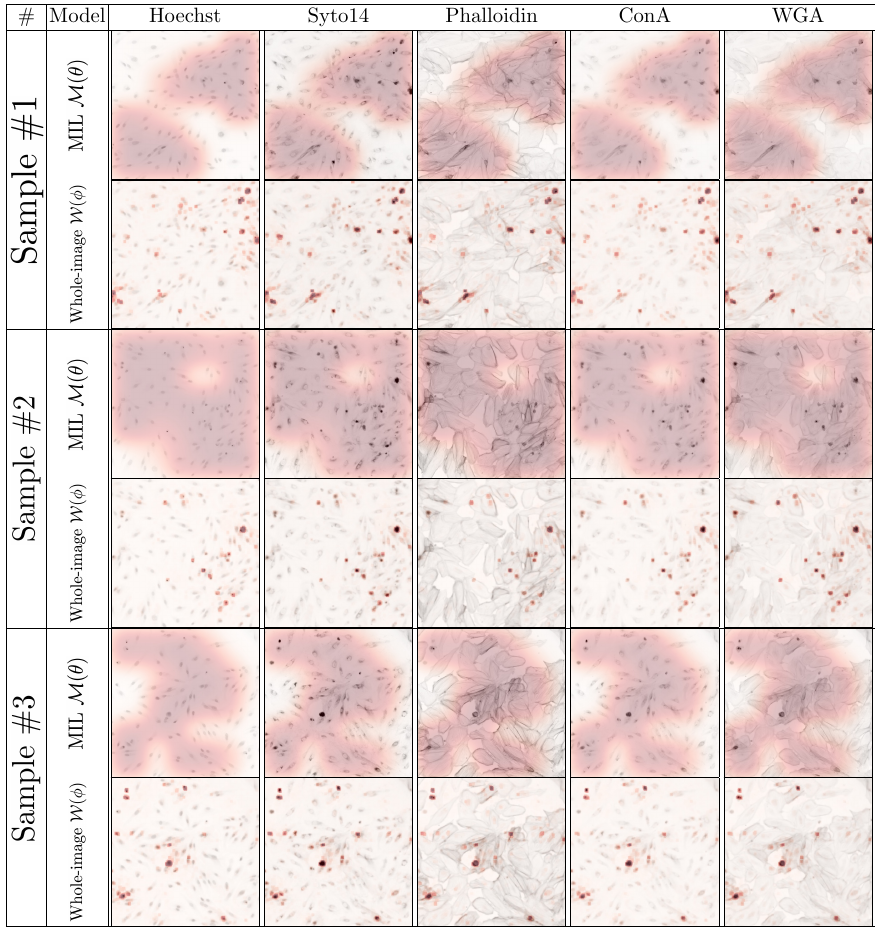}
\caption{{Generated infection maps based on both {MIL model $\mathcal{M}(\theta)$, similar to those in Fig.~5, and whole-image based model $\mathcal{W}(\phi)$} for 3 randomly selected infected samples from the untreated test set. The \textbf{Model} column indicates which model was used in generating the maps. The intensity of the red color shows the probability of the infection on each position based on each model's predictions. For better visualization, each dye image is color-inverted.}}\label{Fig:cpe}
\end{figure*}

We used saliency maps for the whole-image model $\mathcal{W}(\phi)$ predictions and compare them to the infection maps generated by the MIL model $\mathcal{M}(\theta)$. We used \textbf{SHAP} (SHapley Additive exPlanations) to generate saliency maps and estimate the contribution of each input for a specific prediction \cite{lundberg2017unified}. Broadly speaking, SHAP assigns each input feature an importance value for a particular prediction. SHAP is a gradient method to compute SHAP values, which are based on Shapley values proposed in cooperative game theory. Gradient SHAP adds Gaussian noise to each input sample multiple times, selects a random point along the path between baseline and input, and computes the gradient of outputs with respect to those selected random points. The final SHAP values represent the expected value of gradients times the difference between the input and the baselines. We used the CAPTUM library for PyTorch which contains the implementations for many interpretability algorithms such as SHAP \cite{kokhlikyan2020captum}. The results for three examples are shown in Fig~\ref{Fig:cpe}. {For comparison, the generated saliency maps based on the MIL model $\mathcal{M}(\theta)$ are also presented. We can see that the maps generated by both models are mostly in spatial agreement, however, the signal of the maps corresponding to the  whole-image model $\mathcal{W}(\phi)$ are weaker and sparser compared to those of the MIL model $\mathcal{M}(\theta)$.}

\begin{table}[h!]
\centering
\begin{tabular}{c l}
Notation & Description \\
\hline 
$\mathcal{D}$ & Training set in form of $(\mathbf{X},\mathbf{Y})$\\
$n$ & Number of data points in $\mathcal{D}$  \\  
$x_i$ & Fluorescence microscopy sample image\\
$y_i$ & Sample-level label of $x_i$ \\ 
$N$ & Number of patches in each $x_i$\\
$x_i^{p_{j}}$ & j-th patch of $x_i$\\ 
$y_{i}^{p_j}$ & Patch-level label of $x_i^{p_{j}}$\\
${\mu}_i^{p_j}$  & Predicted patch-level label of $x_i^{p_{j}}$\\
$\mathbb{P}[A]$ & Probability of the event $A$ \\
 $M_{i}$ & Set of all predicted patch-level labels for $x_i$ \\
 $\mathcal{K}_{i}^k$ & Set of $k$-top patches with highest ${\mu}_i^{p_j}$ for $x_i$ \\
  $m(S,r)$ & $r$-th greatest element in the set $S$ \\
  $\mathcal{L}(\cdot)$ & Binary cross entropy loss \\
  $w^+$, $w^-$ & Class weights \\
  $\hat{y}_i$ & Predicted sample-level label for $x_i$\\
  $\eta$ & Cut-off threshold for sample classification \\
  $\mathbb{I}(\cdot)$ & Indicator function\\
 $A_i$ & Calculated infection map for sample $x_i$\\
  $\mathcal{O}_{x_i}^{(l,m)}$ & Set of patches overlapping at pixel $x_i^{(l,m)}$ \\
$z_i$ & Infection probability of $x_i$\\
$t_i$ & i-th treatment \\
$c_j$ & j-th concentration \\
$\mathcal{T}_{t_i}^{c_j}$ & Set of all replicates treated with ${t_i}$ at ${c_j}$\\
$\mathcal{C}_{t_i}$ & Set of all concentrations of $t_i$ \\
$e_{t_i}^{c_j}$ & Dose-dependent efficacy score for ${t_i}$ at ${c_j}$\\
 $\beta_{t_i}^{c_j}$ & Median point estimator for $z_i$'s of $\mathcal{T}_{t_i}^{c_j}$\\
CI$(\cdot)$ & Confidence interval an estimation\\
sup & Supremum (least upper bound) \\
$T(\cdot)$ & Descriptive statistic\\
$\zeta$ & Cut-off threshold for effective treatment\\
$\Omega(\omega)$ & Model $\Omega$ with learned weights $\omega$ \\
$\mathcal{E}_{\Omega(\omega)}$ & Set of effective identified treatments by $\Omega(\omega)$\\
$\mathcal{W}(\phi)$ & Whole-image based model \\
$\mathcal{V}(\psi)$ & Patch-based model \\
$\mathcal{M}(\theta)$  & MIL deep neural network model\\

\hline
\end{tabular}
\caption{Table of Notations}\label{notations-tab:1}
\end{table}


\clearpage
\bibliographystyle{IEEEtran}
\bibliography{bib}

\end{document}